\documentclass[10pt,twocolumn,letterpaper]{article}

\usepackage{iccv}
\usepackage{times}
\usepackage{epsfig}
\usepackage{graphicx}
\usepackage{amsmath}
\usepackage{amssymb}
\usepackage{amsmath,amsfonts,bm}
\usepackage{multirow}
\usepackage{subfigure}
\usepackage{booktabs}
\usepackage{array}
\usepackage{bbding}
\usepackage{algorithm, algorithmic}
\usepackage[flushleft]{threeparttable}
\usepackage[accsupp]{axessibility}
\usepackage{enumitem,kantlipsum}
\def\bs{\bm}

\DeclareMathOperator*{\argmin}{argmin}
\usepackage[pagebackref=true,breaklinks=true,letterpaper=true,colorlinks,bookmarks=false]{hyperref}

\iccvfinalcopy % *** Uncomment this line for the final submission

\def\iccvPaperID{9222}   % *** Enter the ICCV Paper ID here

\ificcvfinal\fi

\begin{document}

%%%%%%%%% TITLE
\title{JEM++: Improved Techniques for Training JEM}

\author{Xiulong Yang, Shihao Ji\\
Department of Computer Science \\
Georgia State University\\
{\tt\small \{xyang22,sji\}@gsu.edu}
% \and
% The first line of institution2 address\\
}

\maketitle
\ificcvfinal\thispagestyle{empty}\fi

%%%%%%%%% ABSTRACT
\begin{abstract}

Joint Energy-based Model (JEM)~\cite{jem} is a recently proposed hybrid model that retains strong discriminative power of modern CNN classifiers, while generating samples rivaling the quality of GAN-based approaches. In this paper, we propose a variety of new training procedures and architecture features to improve JEM's accuracy, training stability, and speed altogether. 1) We propose a proximal SGLD to generate samples in the proximity of samples from previous step, which improves the stability. 2) We further treat the approximate maximum likelihood learning of EBM as a multi-step differential game, and extend the YOPO framework~\cite{zhang2019you} to cut out redundant calculations during backpropagation, which accelerates the training substantially. 3) Rather than initializing SGLD chain from random noise, we introduce a new informative initialization that samples from a distribution estimated from training data. 4) This informative initialization allows us to enable batch normalization in JEM, which further releases the power of modern CNN architectures for hybrid modeling.\footnote{Code: \url{https://github.com/sndnyang/JEMPP}}
\end{abstract}

\section{Introduction}

Deep neural networks (DNNs) have made significant breakthroughs in various discriminative tasks and generative tasks, including image classification, object detection, and high-quality image and text generation~\cite{Krizhevsky2012, resnet16, biggan, gpt3}. However, prior works on discriminative models and generative models are largely separated. Even though a few researches (e.g.,~\cite{ssl09,dempster1977maximum}) have shown that generative training is beneficial to discriminative models, most recent works on generative models focus primarily on qualitative sample quality~\cite{biggan2018largeGAN,santurkar2019image,vahdat2021NVAE}, and the discriminative performances of state-of-the-art generative models are still far behind discriminative ones~\cite{behrmann2018invertible, chen2019residual, du2019implicit}. 

Among different discriminative and generative models, energy-based models (EBMs)~\cite{lecun2006tutorial} are an appealing class of probabilistic models, which can be viewed as hybrid models with both discriminative and generative powers~\cite{jem}. Compared to the popular generative models, such as VAE~\cite{VAE} and GAN~\cite{GAN}, which train explicit functions to generate samples, EBMs only need to train a single network with a set of shared features for discriminative tasks and generative tasks, and exploit implicit sampling for generation. Since an EBM is the only object that needs to be trained, it generally achieves a higher simplicity and stability than approaches that use multiple networks. Hence, there is a great interest recently in encompassing the generative capabilities into discriminative models without sacrificing their discriminative powers. Specifically, a series of recent works propose to train a CNN as an EBM for image classification and generation~\cite{xie2016theory, HanNijFan19, du2019implicit, jem}. Among them, JEM~\cite{jem} is one of the most representative ones, which reinterprets the modern CNN classifier (e.g., Wide-ResNet~\cite{wideresnet16}) as an EBM for image generation and achieves impressive performances in image classification and image generation simultaneously. JEM demonstrates the potential of EBMs in hybrid modeling and ignites a series of follow-up works~\cite{ZhaJacGra20, grathwohl2020SteinEBM, Gao2020FlowEBM, nomcmc}.

However, training EBMs is still a challenging task. As shown in Table~\ref{table:ebm_approaches}, existing methods demonstrate a great deal of tradeoffs among different algorithmic features in the quest of improved training algorithms. Most of the works~\cite{nijkamp2019learning, du2019implicit, jem} adopt the SGLD sampling~\cite{welling2011bayesian} to train EBMs, where $K$ sweeps of forward and backward propagations are required in each sampling step. These training methods can be prolonged with a large $K$, preventing them from long training procedures required by large-scale datasets. In addition, SGLD can be precarious and easily diverged, which further hinders the prevalence of EBMs. To avoid the long sampling process of SGLD, recent works introduce auxiliary models~\cite{ebmvae,XieLuGao20,nomcmc} or use special architectures~\cite{grathwohl2020SteinEBM,vincent2011connection} to amortize the SGLD sampling or improve its stability. Given the architectural simplicity of the SGLD-based methods, especially JEM~\cite{jem}, we ask the following question: Is it possible to develop new training methods of JEM to reduce the number of sampling steps required by SGLD while improving its training stability?

\begin{table*}[ht!]
\caption{Characteristics of different EBM training methods.}
\label{table:ebm_approaches}
\begin{center}
\begin{threeparttable}
\begin{tabular}{lcccccc}
\toprule
Training Method & Fast & Stable & High
dim & No aux. model & Unrestricted arch & Approx. likelihood  \\
\midrule
SGLD-based~\cite{nijkamp2019learning, du2019implicit, jem} & \XSolidBrush & \XSolidBrush  & \Checkmark & \Checkmark & \Checkmark & \Checkmark  \\
Score Matching~\cite{vincent2011connection,DDPM}      & \Checkmark & \XSolidBrush & \Checkmark & \Checkmark & \XSolidBrush & \XSolidBrush \\
Noise Contrastive~\cite{Gao2020FlowEBM,gutmann2010noise}  & \Checkmark & \Checkmark & \XSolidBrush & \XSolidBrush & \Checkmark & \XSolidBrush \\
Regularized Generator~\cite{ebmvae,nomcmc}              & \Checkmark & \Checkmark & \Checkmark & \XSolidBrush & \Checkmark & \Checkmark    \\
\midrule
JEM++ (ours)        & $\uparrow$ & $\uparrow$ & \Checkmark & \Checkmark & \Checkmark & \Checkmark  \\
\bottomrule
\end{tabular}
\end{threeparttable}
\vspace{-15px}
\end{center}
\end{table*}

In this paper, we introduce a variety of training procedures and architecture features to improve JEM's accuracy, training stability, and speed altogether. After a thorough investigation on JEM, we find that JEM sometimes generates abnormal images containing pixels with extreme values beyond a reasonable range. This motivates us to constrain the SGLD sampling by projecting samples to an $L_p$-norm ball of previous samples. Secondly, JEM does not support modern architecture features such as batch norm~\cite{batchnorm15}\footnote{Although the authors stated they have been able to train JEM with batch norm successfully, no details are disclosed in their paper or code.}. We find that a huge statistic gap between the initial noisy samples of SGLD and real data incurs the training difficulty of JEM when batch norm is enabled. Hence, we introduce a new informative initialization that closes the gap between initial samples and real data. Moreover, we find that batch-norm-enabled JEM supports a larger learning rate, which further increases the convergence rate of JEM. Finally, we extend YOPO~\cite{zhang2019you}, a general framework for PGD~\cite{madry2018towards} acceleration, to the maximum likelihood learning of EBM and speed up the training of JEM even further. Our main contributions are summarized as follows:
\begin{enumerate}[leftmargin=*, topsep=1pt, partopsep=1pt, itemsep=1pt,parsep=1pt]
\item We propose a proximal SGLD to generate samples in the proximity of samples from previous step, which improves the stability of JEM.
\item We further treat the approximate maximum likelihood learning of EBM as a multi-step differential game, which can be accelerated by cutting out redundant calculations during backpropagation, while retaining the overall predictive performance.
\item We introduce a new informative initialization to initialize the SGLD chain, which stabilizes the training further and accelerates the convergence rate of SGLD sampling.
\item This new informative initialization also enables batch norm to train JEM successfully and release the power of modern CNN architectures. What's more, with the informative initialization and batch norm, JEM++ can be optimized with a large learning rate, while JEM fails to. 
\item JEM++ matches or outperforms prior state-of-the-art hybrid models on discriminative and generative tasks, while enjoying improved stability and training speed over the original JEM.
\end{enumerate}

\section{Energy-Based Models}

Energy-based models (EBMs)~\cite{lecun2006tutorial} define an energy function that assigns low energy values to samples drawn from data distribution and high values otherwise, such that any probability density $p_{\bs{\theta}}(\bs{x})$ can be expressed via a Boltzmann distribution as
\begin{equation}\label{eq:ebm_define}
  p_{\bs{\theta}}(\bs{x})=\frac{\exp \left(-E_{\bs{\theta}}(\bs{x})\right)}{Z(\bs{\theta})},
\end{equation}
where $E_{\bs{\theta}}(\bs{x})$ is an energy function that maps each input $\bs{x}\in\mathcal{X}$ to a scalar, and $Z(\bs{\theta})$ is the normalizing constant (also known as the partition function) such that $p_{\bs{\theta}}(\bs{x})$ is a valid density function. 

The key challenge of training EBMs lies in estimating the partition function $Z(\bs{\theta})$, which is notoriously intractable. The standard maximum likelihood estimation of parameters $\bs{\theta}$ is not straightforward either, and a number of sampling-based approaches have been proposed to approximate it effectively. Specifically, the derivative of the log-likelihood of a single sample $\bs{x}\in\mathcal{X}$ w.r.t. $\bs{\theta}$ can be expressed as
\begin{align}\label{eq:ml}
  \frac{\partial\log p_{\bs{\theta}}(\bs{x})}{\partial\bs{\theta}}=\mathbb{E}_{p_{\bs{\theta}}(\bs{x}')}\frac{\partial E_{\bs{\theta}}(\bs{x}')}{\partial\bs{\theta}}-\frac{\partial E_{\bs{\theta}}(\bs{x})}{\partial\bs{\theta}},
\end{align}
where the expectation is over the density function $p_{\bs{\theta}}(\bs{x}')$, sampling from which is challenging due to the intractable $Z(\bs{\theta})$. Therefore, MCMC and Gibbs sampling~\cite{hinton2002cd} have been proposed previously to estimate the expectation efficiently. To speed up the mixing for effective sampling, recently Stochastic Gradient Langevin Dynamics (SGLD)~\cite{welling2011bayesian} has been employed to train EBMs by using the gradient information~\cite{nijkamp2019learning,du2019implicit,jem}.  Specifically, to sample from $p_{\bs{\theta}}(\bs{x})$, SGLD follows
\begin{align}\label{eq:sgld}
    &\bs{x}^0\sim p_0(\bs{x}),\nonumber\\
    &\bs{x}^{t+1} = \bs{x}^t-\frac{\alpha}{2} \frac{\partial
    E_{\bs{\theta}}(\bs{x}^t)}{\partial \bs{x}^t} + \alpha\epsilon^t, \;\; \epsilon^t \sim \mathcal{N} (0,1),
\end{align}
where $p_0(\bs{x})$ is typically a uniform distribution over $[-1,1]$, whose samples are refined via a noisy gradient decent with step-size $\alpha$ over a SGLD chain. 

Prior works~\cite{du2019implicit, jem, nijkamp2019anatomy} have investigated the effect of hyper-parameters in SGLD sampling in terms of stability and speed, and showed that the SGLD-based approaches suffer from poor stability and computational challenges from sequential sampling at every iteration. Specifically, Nijkamp et al.~\cite{nijkamp2019anatomy} find that the noise term in SGLD is not important, and including a noise of low variance appears to improve synthesis quality. What's more, for unnormalized densities, it's desirable to generate samples from SGLD chain after it converges. This requires the step-size $\alpha$ to decay with a polynomial schedule and an infinite number of sampling steps, which is not realistic in practical applications. Instead, JEM~\cite{jem} uses a constant step-size $\alpha$ during sampling and approximates the samples with a sampler that runs only for a finite number of steps. To improve the sampling stability, the model would require to quadruple the number of SGLD steps, which greatly increases the run-time.

\section{JEM++: The Improved Training of JEM}\label{sec:method}

We first give a brief introduction of JEM~\cite{jem} and then discuss a variety of new training procedures to improve its accuracy, stability and speed.

Joint Energy-based Models (JEM)~\cite{jem} reinterprets modern CNN classifiers as EBMs. Considering a CNN classifier of parameters $\bs{\theta}$, given an input $\bs{x}$ the classifier first maps the input to a vector of $C$ real-valued numbers (or logits): $f_{\bs{\theta}}(\bs{x})[y], \forall y\in [1,\cdots,C]$, where $C$ is the number of classes; the logits are then normalized via the softmax function to yield a probability vector: $p_{\bs{\theta}}(y|\bs{x})=e^{f_{\bs{\theta}}(\bs{x})[y]} / \sum_{y^{\prime}} e^{f_{\bs{\theta}}(\bs{x})\left[y^{\prime}\right]}$. Interestingly, the same vector of logits $f_{\bs{\theta}}(\bs{x})[y]$ can also be used to define an EBM for the joint density: $p_{\bs{\theta}}(\bs{x}, y)=e^{f_{\bs{\theta}}(\bs{x})[y]} / Z(\bs{\theta})$, where $Z(\bs{\theta})$ is an unknown normalizing constant (regardless of $\bs{x}$ or $y$). Then a marginal density of $\bs{x}$ can be achieved by marginalizing the joint  density as: $p_{\bs{\theta}}(\bs{x})=\sum_{y} p_{\bs{\theta}}(\bs{x}, y) = \sum_{y} e^{f_{\bs{\theta}}(\bs{x})\left[y\right]} / Z(\bs{\theta})$. Comparing this density with Eq.~\ref{eq:ebm_define}, it is readily to show that the corresponding energy function of $\bs{x}$ is defined as 
\begin{equation}\label{eq:jem_ex}
    E_{\bs{\theta}} (\bs{x})\!=\!-\log\! \sum_{y}\!e^{f_{\bs{\theta}}(\bs{x})\left[y\right]}\!=\!-\text{LSE}( f_{\bs{\theta}}(\bs{x})),
\end{equation}
where $\text{LSE}(\cdot)$ denotes the Log-Sum-Exp function.

To optimize the model parameter $\bs{\theta}$, JEM proposes to maximize the joint density function $p_{\bs{\theta}}(\bs{x},y)$, which can be factorized as:
\begin{equation}\label{eq:jem_loss}
  \log p_{\bs{\theta}}(\bs{x}, y) = \log p_{\bs{\theta}}(y|\bs{x}) + \log p_{\bs{\theta}}(\bs{x}),
\end{equation}
where the first term is the conventional cross-entropy objective for classification, and the second term can be optimized by the maximum likelihood learning of EBM as shown in Eq.~\ref{eq:ml} with the SGLD sampling defined in (\ref{eq:sgld}). In the paper, we follow the same objective function of JEM and focus on how to improve the stability of SGLD sampling as well as accelerate the maximum likelihood learning of EBM.

\subsection{Training EBM as a Minimax Optimization}\label{sec:minimax}

In practice, when we employ the maximum likelihood estimate of model parameters $\bs{\theta}$ with Eq.~\ref{eq:ml}, a minibatch of $B$ samples $\{\bs{x}_1,\bs{x}_2,\cdots,\bs{x}_B\}\sim p_{\bs{\theta}}(\bs{x})$ and a minibatch of $B$ real data samples $\{\bs{x}^r_1,\bs{x}^r_2,\cdots,\bs{x}^r_B\}\sim \mathcal{X}$ are used. To avoid notational clutter, we assume $B=1$ in the rest of the paper, but the results are readily extended to $B>1$.

Similar to Nijkamp et al.~\cite{nijkamp2019anatomy} who have found the insignificance of noise term in the SGLD sampling (\ref{eq:sgld}), our empirical study also confirms this observation. Thus, we ignore the noise term in Eq.~\ref{eq:sgld} and treat it as an artifact that generates some stochasticity in the sampling process to facilitate the optimization. Under this assumption, the SGLD sampling (\ref{eq:sgld}) can be reinterpreted approximately as an SGD iteration, with a learning rate of $\alpha/2$, initialized from a random sample of $p_0(\bs{x})$. Assume the convergence can be achieved, the objective of the SGLD sampling (\ref{eq:sgld}) is to solve the following optimization problem approximately~\footnote{The entire pipeline is still a stochastic sampler since samples are generated by running finite-length stochastic gradient decent with random initialization.}
\begin{align}\label{eq:minx}
    \bs{x}^*=\argmin_{\bs{x}} E_{\bs{\theta}}(\bs{x}).
\end{align}
Therefore, the maximum likelihood learning of EBM with Eq.~\ref{eq:ml} is to approximately solve the following minimax game
\begin{align}\label{eq:minimax}
    \max_{\bs{\theta}}\left[\min_{\bs{x}}E_{\bs{\theta}}(\bs{x})-E_{\bs{\theta}}(\bs{x}^r)\right].
\end{align}

To have a robust convergence behavior, we can solve the inner minimization problem of (\ref{eq:minimax}) by using the Proximal Point Method~\cite{proximal}. We can further treat the minimax optimization problem (\ref{eq:minimax}) as a multi-step differential game and extend YOPO~\cite{zhang2019you}, a general framework of accelerating PGD, to speed up the training of EBM. Next, we describe these new training procedures in details.

\subsection{Proximal SGLD}

Prior works on EBMs reveal the tradeoff between training stability and computational time of SGLD-based approaches~\cite{nijkamp2019learning,du2019implicit, jem}. However, the cause of instability of SGLD-based EBMs is still under investigation. Empirically, we observe that upon the divergence of EBM, SGLD generates abnormal samples with extreme values that have a severe negative impact on model parameter update. Hence, we introduce our first improvement to stabilize the inner minimization problem with a proximal SGLD.

Proximal point methods are widely used in optimization~\cite{Roc76Proximal,proximal}. To solve the inner minimization problem of (\ref{eq:minimax}), the algorithm generates a sequence $\{\bs{x}^t\}_{t=1,2,\cdots}$ by the following proximal point iteration:
\begin{align}\label{eq:ppi}
    \bs{x}^{t+1} = \argmin_{\bs{x}}E_{\bs{\theta}}(\bs{x}) \;s.t.\;||\bs{x}-\bs{x}^t||_p<\varepsilon,
\end{align}
which solves a constrained minimization problem at each iteration $t$, i.e., the current solution should be in the proximity of previous one, measured by an $L_p$ norm. Compared with the standard SGD iteration, the proximal point iteration has a robust convergence behavior. Moreover, even if the proximal operator defined in Eq.~\ref{eq:ppi} is not exactly minimized in each iteration, it still has a stronger convergence guarantee than standard SGD, giving rise to the inexact proximal point method~\cite{Roc76Proximal}. Thus, if we solve each minimization problem (\ref{eq:ppi}) inexactly with one step of SGD, we obtain an inexact proximal point iteration
\begin{align}\label{eq:ppi_sgd}
    \bs{x}^{t+1} = \bs{x}^{t} - \frac{\alpha}{2}L_p(\nabla_{\bs{x}} E_{\bs{\theta}}(\bs{x}^t), \varepsilon),
\end{align}
where $L_p(\cdot,\varepsilon)$ projects the gradient to an $L_p$-norm ball of a radius $\varepsilon$. Empirically, we find the $L_{\infty}$-norm works well across different architectures and datasets. Hence, we only consider the $L_{\infty}$-norm in the rest of the paper. With an $L_\infty$-norm, Eq.~\ref{eq:ppi_sgd} can be rewritten as
\begin{align}\label{eq:psgld}
    \bs{x}^{t+1} = \bs{x}^{t} - \frac{\alpha}{2}\text{clamp}(\nabla_{\bs{x}} E_{\bs{\theta}}(\bs{x}^t), \varepsilon) + \alpha\epsilon^t,
\end{align}
where the $\text{clamp}(\cdot,\varepsilon)$ operator clamps the gradient in the range of $[-\varepsilon,\varepsilon]$. Note that to incorporate stochasticity into the inexact proximal point iteration, we add the noise term back to Eq.~\ref{eq:psgld}, which resembles the original SGLD sampling (\ref{eq:sgld}) but with a gradient clamping operator used to enforce the proximity constraint.

\subsection{Training EBM as a Differential Game}

As discussed in Section~\ref{sec:minimax}, the maximum likelihood learning of EBM (\ref{eq:minimax}) solves a minimax game approximately. This objective has a close relationship to adversarial training with the PGD attack~\cite{madry2018towards}. Hence, we can extend methods for accelerating adversarial training to EBM and reduce the computational complexity of multi-step SGLD. 

Inspired by Pontryagin’s Maximum Principle~\cite{PMP87}, a general framework in optimal control, Zhang et al.~\cite{zhang2019you} propose an optimization method called YOPO (\textbf{Y}ou-\textbf{P}ropogate \textbf{O}nly \textbf{O}nce) to accelerate multi-step adversarial training such as PGD. The key factor in YOPO is that the adversarial perturbation is only coupled with the first layer's weights in a neural network. Then YOPO can decouple the adversary update from training of network parameters, and reduce the total number of full forward and backward propagations to only one in each group of adversary updates. 

Similarly, we can extend YOPO to the maximum likelihood learning of EBM because the objective (\ref{eq:minimax}) can also be treated as a multi-step differential game and the sampled image $\bs{x}$ from proximal SGLD (\ref{eq:psgld}) is only coupled with the first layer's weights. By inserting the energy function (\ref{eq:jem_ex}) into (\ref{eq:minimax}), we can rewrite the minimax objective as:
\begin{align}
\max _{\bs{\theta}} & \left[ \min _{ \bs{x} }
    -\text{LSE}\left(g_{\tilde{\bs{\theta}}}\left(f_{0}\left(\bs{x}, \bs{\theta}_{0}\right)\right)\right) -E_{\bs{\theta}}(x^r)  \right] 
\end{align}
where $f_0$ denotes the first layer of a CNN-based EBM, $g_{\tilde{\bs{\theta}}}=f_{T-1}^{\bs{\theta}_{T-1}} \circ f_{T-2}^{\bs{\theta}_{T-2}} \circ \cdots f_{1}^{\bs{\theta}_{1}}$ denotes the network without the first layer, such that $f_{\bs{\theta}}(\bs{x})=g_{\tilde{\bs{\theta}}}\left(f_{0}\left(\bs{x}, \bs{\theta}_{0}\right)\right)$. Given a sample $\bs{x}$, the gradient of energy function (\ref{eq:jem_ex}) can be calculated by chain rule as:
\begin{align}
  \frac{\partial E_{ \bs{\theta} }(\bs{x})}{\partial \bs{x}} = & -\nabla_{g_{\tilde{\bs{\theta}}}} \text{LSE} \left(g_{\tilde{\bs{\theta}}}\left(f_{0}\left(\bs{x}, \bs{\theta}_{0}\right)\right) \right) \nonumber \\
   & \cdot \nabla_{f_0}g_{\tilde{\bs{\theta}}}(f_0(\bs{x}, \bs{\theta}_{0})) \cdot  \nabla_{\bs{x}}f_0(\bs{x}, \bs{\theta}_0).
\end{align}

\begin{figure}[t]
    \centering
    \includegraphics[width=1.0\columnwidth]{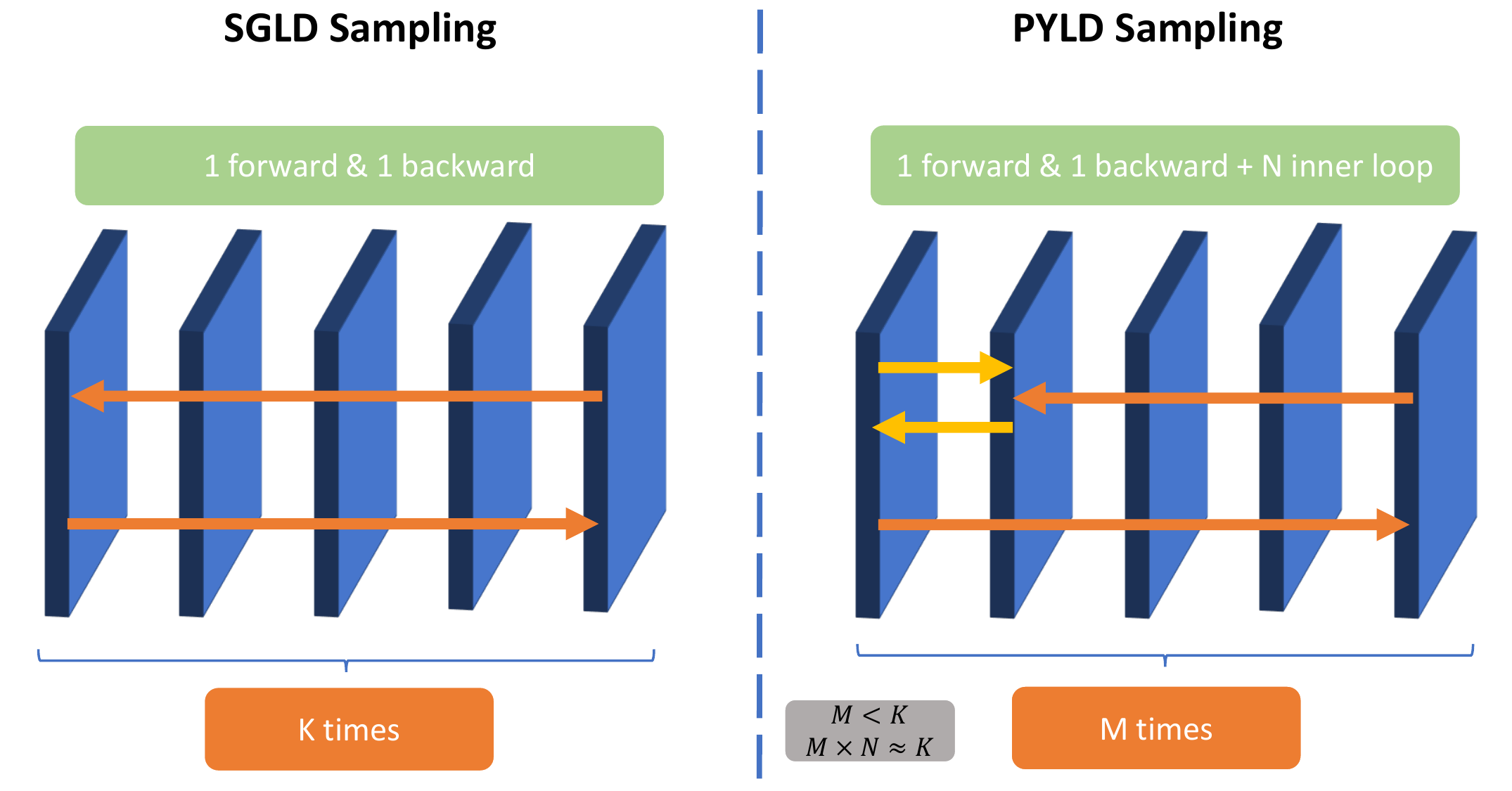}
    \caption{Comparison between SGLD-$K$ sampling and PYLD-$M$-$N$ sampling.}\label{figure:yopo_diagram}
    \vspace{-10px}
\end{figure}

Proximal SGLD (\ref{eq:psgld}) conducts $K$ sweeps of full forward and backward propagations for each update of $\bs{\theta}$. To stabilize the training of EBM, it requires a large $K$, which greatly increases the run-time. To reduce the total number of thorough forward and backward propagations, we follow YOPO and introduce a slack variable:
\begin{align}
p\!=\!-\nabla_{g_{\tilde{\bs{\theta}}}}\text{LSE}\! \left(g_{\tilde{\bs{\theta}}}\left(f_0\!\left(\bs{x}, \bs{\theta}_{0}\right)\right) \right)
   \cdot \nabla_{f_0}g_{\tilde{\bs{\theta}}}\left(f_{0}\!\left(\bs{x},\!\bs{\theta}_{0}\right)\right),
\end{align}
and freeze it as a constant in the inner loop of the sample update. We call our accelerated Proximial SGLD algorithm PYLD-$M$-$N$ (\textbf{P}roximal-\textbf{Y}OPO-SG\textbf{LD}) with $M$ outer loops and $N$ inner loops. Figure~\ref{figure:yopo_diagram} demonstrates a conceptual comparison between SGLD-$K$ and PYLD-$M$-$N$. SGLD-$K$ accesses the data $K$ times requiring $K$ full forward and backward propagations. On the contrary, PYLD-$M$-$N$ accesses the data $M\times N$ times, while only requiring $M$ full forward and backward propagations and a inner loop of $M\times N$ cheap sample updates. Similar to YOPO~\cite{zhang2019you}, when $M\times N\approx K$, PYLD can achieve a similar sample quality as SGLD. But PYLD-$M$-$N$ has the flexibility of increasing $N$ and reducing $M$ to achieve approximately the same level of movement with much less computation cost. We will demonstrate this when we present results.

The pseudo-code of our PYLD is described in Algorithm~\ref{algo:yopo}. For more details of YOPO, we refer the readers to~\cite{zhang2019you}. 

\begin{algorithm}[t]
\caption{PYLD-$M$-$N$ sampling: Given network $g_{\tilde{\bs{\theta}}}$ and $f_{0}$ with $\bs{\theta}_0$, step-size $\alpha$, number of steps $M$ and $N$}
\label{algo:yopo}
\begin{algorithmic}[1]
%\STATE Given each input $\bs{x}_i$  (MAYBE remove $_i$ for brevity,)
\STATE $\bs{x}^0 \sim p_0(\bs{x})$
\FOR{$t \in [0, 1, \cdots, M-1]$}
  %\STATE $\eta^0 = \nabla_{\bs{x}^t} E_{\bs{\theta}}(\bs{x} ^ t)$
  \STATE \% calculate the slack variable
  \STATE $p\!=\!-\nabla_{g_{\tilde{\bs{\theta}}}}\text{LSE}\! \left(g_{\tilde{\bs{\theta}}}\!\left(f_0\!\left(\bs{x}^t,\!\bs{\theta}_0\right)\right)\right)
   \cdot \nabla_{f_0}g_{\tilde{\bs{\theta}}}\!\left(f_{0}\!\left(\bs{x}^t,\!\bs{\theta}_0\right)\right)$
  \STATE $\bs{x}^{t, 0}=\bs{x}^{t}$ %+ \alpha \eta^0$
  %\STATE $\eta^0 = 0$  
  \FOR{$s \in [0, 1, \cdots, N-1]$}
    \STATE $\gamma = \text{clamp}(p \cdot\nabla_{\bs{x}^{t, s}}  f_0\!\left(\bs{x}^{t, s}, \bs{\theta}_0\right), \varepsilon)$
    \STATE $\bs{x}^{t, s+1}=\bs{x}^{t, s} -\alpha/2\cdot\gamma$
    %\STATE $\eta^{s+1} = \eta^{s} + \gamma$
  \ENDFOR
  \STATE $\bs{x}^{t+1}=\bs{x}^{t,N} + \alpha \epsilon^t$%\bs{x}^{t} - \alpha/2\cdot\text{clamp} (\eta^N,  \varepsilon) + \alpha \epsilon^t $
\ENDFOR
% \STATE Calculate the weight update
% \STATE $U=\frac{1}{M} \sum_{j=1}^{M} \nabla_{\bs{\theta}}  E \left(g_{\tilde{\bs{\theta}}}\left(f_{0}\left(\bs{x}^{j}, \bs{\theta}_{0}\right)\right) \right)$
\STATE return $\bs{x}^M$
\end{algorithmic}
\end{algorithm}

\subsection{Informative Initialization}

The initial sampling distribution $p_0(\bs{x})$ also plays an important role in the training of EBM. Nijkamp et al.~\cite{nijkamp2019anatomy} summarize two main types of SGLD initializations for $\bs{x}^0$: non-informative initialization and informative initialization. The former initializes the sample $\bs{x}^0$ from a noise distribution independent to the training data, such as a uniform or Gaussian distribution, while the latter
samples from an approximate distribution close to the data distribution. One typical informative initialization is to use samples from training data directly, as proposed in Contrastive Divergence (CD)~\cite{hinton2002cd}. Based on this, Tieleman~\cite{tieleman2008training} proposes Persistent Contrastive Divergence (PCD) and uses samples from previous learning iteration as the initial samples for the current iteration. In contrast to common wisdom, Nijkamp et al.~\cite{nijkamp2019learning} propose a short-run MCMC sampler which always starts from the random noise distribution such as a uniform distribution. Moreover, to train EBMs, Xie et al.~\cite{xie2016theory} propose another persistent initialization, which combines non-informative and informative initialization and samples short SGLD chains from data samples of previous iterations and occasionally (with a small probability $\rho$) reinitializes the chains from random noise. This is also the sampling approach adopted by IGEBM \cite{du2019implicit} and JEM~\cite{jem}, which maintain a replay buffer of samples from previous iterations and replace a small percentage of samples in the buffer with random noise to train EBMs.

\begin{figure}[t]
    \vspace{-5px}
    \centering
        \includegraphics[width=0.9\columnwidth]{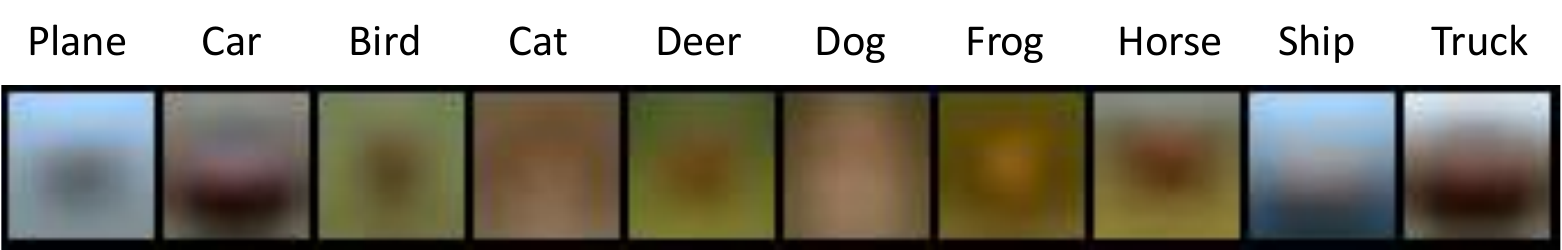}
    \caption{The categorical centers of CIFAR10.}
    \label{fig:informative_init_CIFAR10}
    \vspace{-10px}
\end{figure}

In this paper, we explore informative initialization to initialize the SGLD chain, and use the PCD with a replay buffer. The main difference is that we substitute the random noise samples with samples from a Gaussian mixture distribution estimated from the training dataset. That is, we define the initial sampling distribution as
\begin{align}\label{eq:gmm2}
  p_0(\bs{x})&=\sum\nolimits_y\mathcal\pi_y{N}(\bs{\mu}_y,\bs{\Sigma}_y)\\
  \text{with}\quad \pi_y &= |\mathcal{D}_y|/\sum\nolimits_{y'}|\mathcal{D}_{y'}|,\quad
\bs{\mu}_{y} = \mathbb{E}_{\bs{x} \sim \mathcal{D}_{y}}[\bs{x}], \nonumber \\ 
\bs{\Sigma}_y&=\mathbb{E}_{\bs{x} \sim \mathcal{D}_{y}}\left[\left( \bs{x} - \bs{\mu}_{y}\right)\left( \bs{x} -\bs{\mu}_{y}\right)^{\top}\right],\nonumber
\end{align}
where $\mathcal{D}_{y}$ denotes the set of training samples with label $y$. As an example, Figure~\ref{fig:informative_init_CIFAR10} visualizes the $\{\bs{\mu}_1, \bs{\mu}_2,\cdots, \bs{\mu}_{10}\}$ (categorical centers) estimated from the CIFAR10 training dataset. Similar visualizations on CIFAR100 and SVHN as well as example samples from the informative initialization can be found in the supplementary material.
%\cite{santurkar2019image} proposed a simple initialization with really sufficient information - a multivariate normal distribution fit to the empirical class-conditional distribution. 

The informative initialization brings sufficient information into $\bs{x}^0$ to guide the SGLD chain to converge faster than from a random noise since the initial sample $\bs{x}^0$ is now much closer to the real data manifold. Empirically, we also observe the improved training stability. What's more, the informative initialization allows us to enable batch norm~\cite{batchnorm15}, a modern architecture feature of DNNs, that is excluded by IGEBM and JEM due to the training difficulty introduced by batch norm. 

%prior works~\cite{du2019implicit,jem} fail to enable modern architecture features, such as batch norm~\cite{batchnorm15}, which limit the power of modern CNN architectures. However, when we sample $\bs{x}_0$ from the distribution of natural inputs conditioned on the label $y$, the statistic of $\bs{x}_0$ supports the usage of advanced neural network components to fully release the expressive power of model deep learning architectures as we will explain in the next section.

\subsection{Batch Normalization and Learning Rate}

Batch norm~\cite{batchnorm15} is an essential component in many state-of-the-art CNN architectures. Batch norm normalizes input features by the mean and variance computed within each mini-batch, which mitigates the vanishing gradient issue of training very deep networks and dramatically improves the convergence rate of gradient-based methods. Moreover, batch norm allows a much larger learning rate and mitigates the need of tedious finetuning.

However, state-of-the-art EBMs, such as IGEBM~\cite{du2019implicit} and JEM~\cite{jem}, do not support batch norm. If batch norm is enabled in JEM, the model can neither achieve a high classification accuracy nor generate realistic images. This is because one intrinsic assumption of batch norm is that the input features should come from a single or similar distributions. This normalization behavior could be problematic if the mini-batch contains data from different distributions, therefore resulting in inaccurate statistics estimation. Unfortunately, this might be the case for the original IGEBM and JEM. Apparently, if the initial samples $\bs{x}^0$ are sampled from a uniform or Gaussian distribution as in IGEBM and JEM, $\bs{x}^0$ and real data samples have different underlying distributions, violating the assumption of batch norm.

Similar phenomenon has also been observed by Xie et al.~\cite{AdvBN} who demonstrates the different statistics between clean data and adversarial examples. They show that both clean data accuracy and adversarial robustness can be improved by using two branches of batch norm: one main branch for clean data and one auxiliary branch for adversarial examples. Instead of using two batch norms, we mitigate the training difficulty of batch norm from a different perspective. Since we have the choice of designing sampling distribution $p_0(\bs{x})$, we can use the informative initialization discussed above to enable batch norm in the EBM training. Since the Gaussian mixture distribution (\ref{eq:gmm}) is actually estimated from real training examples, we can close the statistic gap between initial samples of SGLD and real data and enable batch norm in JEM++ successfully. What's more, with the informative initialization and batch norm, JEM++ can also use a much larger learning rate to improve convergence rate even further.

In summary, Algorithm~\ref{algo:1} provides the pseudo-code for JEM++ training, which follows a similar design of JEM~\cite{jem} and IGEBM~\cite{du2019implicit} with a replay buffer. For brevity, only one real sample $(\bs{x}^r,y^r)\sim\mathcal{D}$ and one generated sample $\bs{x}^M\sim p_{\bs{\theta}}(\bs{x})$ are used to optimize the parameter $\bs{\theta}$. It is straightforward to generalize the pseudo-code above to a mini-batch setting, which we use in the experiments.

\begin{algorithm}[t]
\caption{Training JEM++: Given network $f_{\bs{\theta}}$, step-size $\alpha$, replay buffer $\mathbb{B}$, number of steps $M$ and $N$, reinitialization frequency $\rho$, and number of classes $C$}
\label{algo:1}
\begin{algorithmic}[1]
\WHILE{not converged}
\STATE Sample $(\bs{x}^r, y^r)\sim \mathcal{D}$
\STATE Sample $\bs{x}^0 \sim \mathbb{B}$ with probability $1-\rho$, else $\bs{x}^0 \sim \mathcal{N}(\bs{\mu}_{y}, \bs{\Sigma}_{y}), y \sim p(y) = \bs{\pi}$
\STATE Apply PYLD in Algo.~\ref{algo:yopo} to sample $\bs{x}^M$ from $\bs{x}^0$
\STATE Calculate gradient with Eq.~\ref{eq:ml} from $\bs{x}^r$ and $\bs{x}^M$, and gradient of CE loss from $(\bs{x}^r,y^r)$, and update model parameter $\bs{\theta}$
\STATE Add / replace sample $\bs{x}^M$ back to $\mathbb{B}$
\ENDWHILE
\end{algorithmic}
\end{algorithm}
%\vspace{-10pt}

\section{Experiments}
We evaluate the performance of JEM++ on multiple discriminative and generative tasks, including
image classification, image generation, adversarial robustness, calibration of uncertainty, and out-of-distribution (OOD) detection. Since our main goal is to improve JEM's accuracy, training stability and speed, we present these results in the main text and relegate its downstream applications, such as adversarial robustness, calibration and OOD detection, to the supplementary material. For a fair comparison with  JEM~\cite{jem}, our experiments closely follow the settings provided in the source code of JEM\footnote{\url{https://github.com/wgrathwohl/JEM}}. All our experiments are performed with PyTorch on Nvidia RTX GPUs. %Our source code is provided as a part of supplementary materials.

\begin{table}[t]
\caption{Hybrid Modeling Results on CIFAR10. We report JEM++'s performance with different $M$s when $N=5$ is fixed. We also report the per epoch speedup between JEM and JEM++.}
\label{table:hybrid_results}
\vspace{-15pt}
\begin{center}
\begin{threeparttable}
\begin{tabular}{c|c|ccc}
\toprule
Class  & Model  & Acc \% $\uparrow$ & IS$^*$ $\uparrow$ & FID$^*$ $\downarrow$ \\
\midrule
        & Residual Flow~\cite{chen2019residual}  & 70.3 & 3.60  & 46.4 \\
        & Glow~\cite{kingma2018glow}           & 67.6 & 3.92 & 48.9 \\
Single  & IGEBM~\cite{du2019implicit}          & 49.1 & 8.30  & 37.9 \\
Hybrid  & JEM (K=20)~\cite{jem} $1\times$    & 92.9 & \bf{8.76} & 38.4 \\
Model   & JEM++ (M=5)  $2.4\times$      & 91.1 & 7.81 & 37.9  \\
        & JEM++ (M=10) $1.5\times$  & 93.5 & 8.29 & 37.1  \\
        & JEM++ (M=20) $.92 \times$   & \textbf{94.1} & 8.11 & 38.0  \\
\midrule
Reg    & VERA$^\dagger$ ($\alpha$=100) $2.8\times$  & 93.2 & 8.11 & 30.5 \\
Gen.   & VERA~\cite{nomcmc}  ($\alpha$=1)  $2.8\times$  & 76.1 & 8.00 & 27.5 \\
\midrule
Disc.  & WRN w/ BN  & 95.8 & N/A & N/A  \\
\midrule
Gen.   & SNGAN~\cite{miyato2018spectral}  &  N/A & 8.59  & 25.5  \\
       & NCSN~\cite{song2019generative}   &  N/A & 8.91  & 25.3 \\
\bottomrule
\end{tabular}
\begin{tablenotes}
  \scriptsize\item $^{\dagger}$VERA uses an auxiliary generator to amortize the SGLD sampling and reports a $2.8\times$ speedup without much details on how the evaluation is performed. 
  \scriptsize\item $^*$A fair evaluation of IS and FID is challenging as different methods use different ways to measure the image quality. JEM uses an ensemble of models to evaluate its IS and FID, while JEM++ only uses a single model for evaluation. No more details are provided in JEM. Thus, it is difficult to have a fair comparison. 
\end{tablenotes}
\end{threeparttable}
\vspace{-15pt}
\end{center}
\end{table}

%We show the evolution of JEM++ in Figure~\ref{figure:acc_loss_is_fid_all}.

\begin{table}[t]
\caption{Test Accuracy (\%) on SVHN and CIFAR100.}
\label{table:svhn_CIFAR100_acc}
\begin{center}
\begin{threeparttable}
\begin{tabular}{l|cc}
\toprule
Model  & SVHN & CIFAR100 \\
\midrule
Softmax (w/ BN) & 97.0 & 78.9  \\
\hline
VERA~\cite{nomcmc}           & 96.8 & 72.2   \\
JEM (K=20)     & 96.7 & 72.2   \\
JEM++ (M=5)    & 96.7 & 72.0   \\
JEM++ (M=10)   & \textbf{96.9} & \textbf{74.5}   \\
\bottomrule
\end{tabular}
\end{threeparttable}
\vspace{-15pt}
\end{center}
\end{table}

\subsection{Hybrid Modeling}

We train JEM++ on three benchmark datasets: CIFAR10, CIFAR100~\cite{Krizhevsky2012} and SVHN~\cite{svhn11}, and compare it to the state-of-the-art hybrid models, as well as standalone generative and discriminative models. Following the settings of JEM~\cite{jem}, all our experiments are based on the Wide-ResNet architecture~\cite{wideresnet16}, with the details of hyper-parameter settings of JEM++ provided in the supplementary material. It's worth mentioning that applying the SGD optimizer with $lr=0.1$ to JEM++ achieves better accuracy than the default setting of JEM using Adam with $lr=0.0001$\footnote{JEM cannot use a learning rate larger than 0.0001. Otherwise, it is extremely unstable and diverges easily at early epochs.}. %, while retaining comparable quality of generated images. 
To evaluate the quality of generated images, we adopt the Inception Score (IS)~\cite{imprgan16} and Fr\'{e}chet Inception Distance (FID)~\cite{heusel2017gans}.  

\begin{figure}[t]
\vspace{-5pt}
    \centering
    \subfigure[Unconditional Samples]{
        \includegraphics[width=0.45\columnwidth]{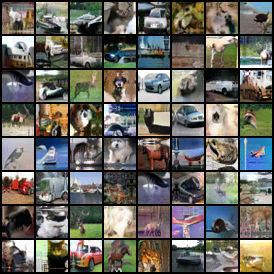}
        \label{figure:CIFAR10_un_sample}
    }
    \subfigure[Class-conditional Samples]{
        \includegraphics[width=0.45\columnwidth]{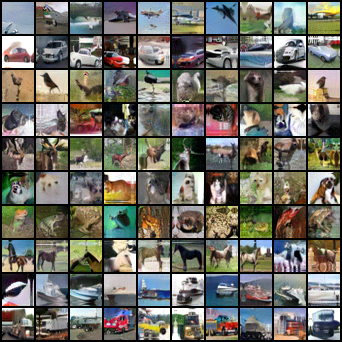}
        \label{figure:CIFAR10_con_sample}
    }
    \caption{JEM++ generated CIFAR10 samples.}
    \label{figure:CIFAR10_samples}
    \vspace{-10px}
\end{figure}

The results on CIFAR10, CIFAR100 and SVHN are reported in Table~\ref{table:hybrid_results} and~\ref{table:svhn_CIFAR100_acc}, respectively. It can be observed that JEM++ ($M\!=\!10$) outperforms JEM and other single-network hybrid models in terms of accuracy (93.5\%), FID score (37.1) and per epoch speedup (1.5$\times$), while being slightly worse in IS score. Since no IS and FID scores are commonly reported on SVHN and CIFAR100, we present the classification accuracy and generated samples on these two benchmarks. Our JEM++ ($M\!\!=\!\!10$) model achieves an accuracy of 96.9\% and 74.5\% on SVHN and CIFAR100, respectively, outperforming JEM by notable margins. Example images generated by JEM++ for CIFAR10, SVHN and CIFAR100 are shown in Figure~\ref{figure:CIFAR10_samples} and~\ref{figure:svhn_cifar100_samples}, respectively. Additional JEM++ generated images can be found in the supplementary material. 

We also investigated JEM++'s performances on several downstream applications, including adversarial robustness, calibration of uncertainty, and OOD detection, where JEM++ achieves improved performances over the original JEM in most of the cases. Due to page limit, the details are relegated to the supplementary material.

\begin{figure}[t]
\vspace{-5pt}
    \centering
    \subfigure[SVHN]{
        \includegraphics[width=0.45\columnwidth]{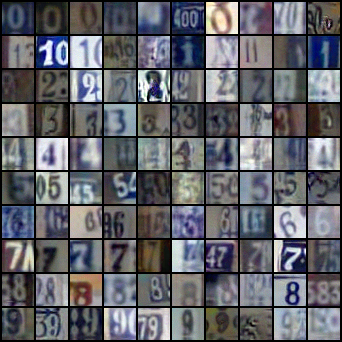}
        \label{figure:svhn_con_sample}
    }
    \subfigure[CIFAR100]{
        \includegraphics[width=0.45\columnwidth]{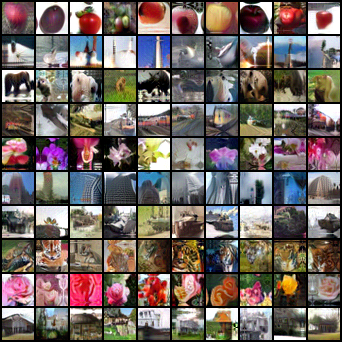}
        \label{figure:CIFAR100_con_sample}
    }
    \vspace{-2pt}
    \caption{JEM++ generated class-conditional samples of SVHN and CIFAR100. Each row corresponds to one class.}
    \label{figure:svhn_cifar100_samples}
    \vspace{-10pt}
\end{figure}

\subsection{Training Stability and Speed}

% JEM++(M=5) is even more stable than JEM(K=20)
% JEM++(M=20) is (nearly) fully stabilized.

The main limitation of the SGLD-based training is the tradeoff between training time and stability. The more SGLD sampling steps are used, the more stable and better performance EBMs can achieve. In this section, we evaluate JEM and JEM++ in terms of training stability and speed.

%Hence, we can easily get that the theoretical speedup of our JEM++ is $(K+1) / (M+1)$\footnote{$K+1$ or $M+1$, because of the one pass of $p_{\bs{\theta}}(y| \bs{x})$} and the cost of inner loop $N \neq 0$ is negligible. 

\begin{figure}[t]
    \centering       
    \includegraphics[width=0.9\columnwidth]{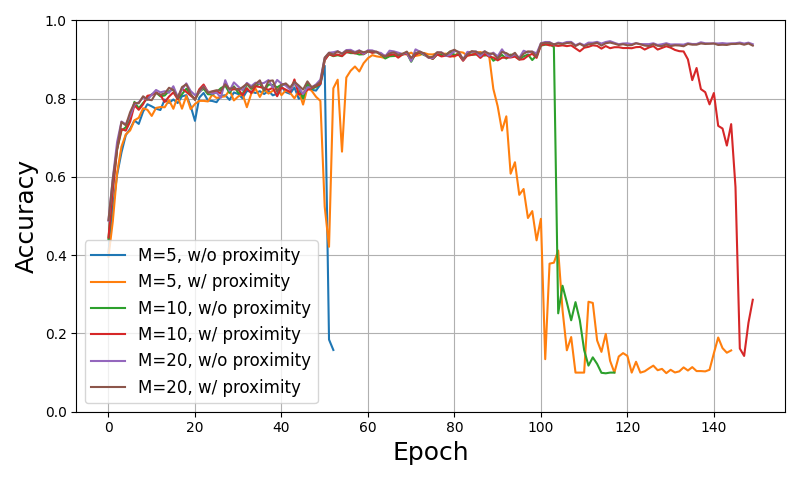}
    \vspace{-5pt}
    \caption{The learning curves of JEM++ trained on CIFAR10 with different configurations: (1) Number of steps $M$ with $N=5$, and (2) the proximity constraint. The official JEM uses $K=20$, but it regularly diverges before 60 epochs.}
    \label{figure:instability_curve}
    \vspace{-15pt}
\end{figure}

We first compare the training stability of JEM and JEM++. From our empirical study, the official JEM ($K$-step SGLD with $K=20$) suffers from training instability, i.e., it regularly diverges before 60 epochs. Prior works~\cite{du2019implicit,jem}, including JEM, fail to find a reasonably small $K$ to completely stabilize the training of EBMs, and thus rely on checkpoints to resume the training when divergences occur. Figure~\ref{figure:instability_curve} shows the learning curves of JEM++ trained on CIFAR10 with different configurations. As can be seen, JEM++ is much more stable and does not diverge when $M\!=\!20$. What's more, JEM++ with $M\!=\!10$ can achieve high stability; even JEM++ with $M\!=\!5$ is more stable than JEM with $K\!=\!20$. As discussed in Section~\ref{sec:method}, the informative initialization improves JEM's stability because the initial samples $\bs{x}^0$ of SGLD are now close to the real data manifold. Hence, the sampling process requires fewer steps to reach the low energy region of the energy function, which we conjure should be much smoother than other regions. In addition, the proximity constraint also improves the stability of JEM++ as demonstrated in Figure~\ref{figure:instability_curve}.

We further compare the training speed between JEM and JEM++ in terms of run-time per epoch. The results are reported in Table~\ref{table:speed}, where we compare JEM and JEM++ trained on CIFAR10 with different configurations of $M$ and $N$. It can be observed that $M$ specifies the total number of forward and backward propagations of PYLD, consuming most of the run-time, while $N$ has a minor impact on the run-time as it specifies the number of inner loops for sample update, which is relatively inexpensive. Therefore, we can increase $N$ and reduce $M$ to achieve approximately the same level of sample quality with much less computation cost. Considering the training stability (Figure~\ref{figure:instability_curve}) and training speed (Table~\ref{table:speed}), $M\!=
\!10$ and $N\!=\!5$ achieves a good balance between the two criteria and therefore is our default configuration of JEM++.

%The official JEM with $K=20$. It is unstable and usually fails before 60 epochs (even less stable than JEM++ with $M=5$

\begin{table}[h!]
\caption{Run-time comparison of JEM and JEM++ on CIFAR10.}
\label{table:speed}
\begin{center}
\vspace{-5pt}
%\begin{threeparttable}
\begin{tabular}{l|cc}
\toprule
Model & Minutes per epoch & Speedup \\
\midrule
JEM             &  30.1  & $1 \times$  \\
\midrule
JEM++, $M=5$    &        &      \\
\qquad $N=5$    &  12.5  & $2.41 \times$  \\
\qquad $N=10$   &  12.6  & $2.39 \times$  \\
\qquad $N=20$   &  13.0  & $2.31 \times$  \\
JEM++, $M=10$   &           \\
\qquad $N=5$    &  20.1  & $1.49 \times$  \\
\qquad $N=10$   &  20.3  & $1.48 \times$  \\
\qquad $N=20$   &  20.4  & $1.47 \times$  \\
JEM++, $M=20$   &           \\
\qquad $N=5$    &  32.5  & $.93 \times$  \\
\qquad $N=10$   &  32.7  & $.92 \times$  \\
\qquad $N=20$   &  32.9  & $.91 \times$  \\
\bottomrule
\end{tabular}
%\begin{tablenotes}
%  \scriptsize\item $^{*}$The official JEM with $K=20$. It is unstable and usually fails before 60 epochs (even less stable than JEM++ with $M=5$).
%\end{tablenotes}
%\end{threeparttable}
\vspace{-20pt}
\end{center}
\end{table}

\subsection{Ablation Study}

JEM++ introduces a variety of new training procedures and architecture features to improve JEM's accuracy, training stability and speed. In this section, we study the effect of different components of JEM++ on the performance of image classification and image generation. Specifically, we conduct the ablation study on CIFAR10 with an exhaustive comparison of different components. We measure the effects of 1) w/o proximity constraint, 2) with Adam optimizer, 3) random initialization with batch norm enabled, and 4) two different types of initialization w/o batch norm.

The results are reported in Table~\ref{table:ablation}. It can be observed that each component contributes to JEM++’s performance positively. The proximity constraint in Proximal SGLD improves both stability and accuracy. Our experiments show that when a smaller $M$ enlarges the instability, the proximity constraint not only helps to stabilize the training, but also improves the accuracy of the trained models. The informative initialization also takes a significant role in JEM++, which enables both batch norm and the use of SGD with larger learning rates. When batch norm is enabled in JEM, we find that it can neither achieve a high classification accuracy nor generate realistic images. On the other hand, JEM without batch norm can achieve decent classification accuracy and generate quality images, but it's precarious and easily diverged at early epochs. The informative initialization itself w/o batch norm is still beneficial to stabilize the training, as manifested by the improved classification accuracy and image quality. It's worth mentioning that when batch norm is disabled, only Adam~\cite{adam2014} with a small learning rate no greater than 0.0001 yields a stable training. However, when batch norm is enabled, the SGD optimizer with a much larger learning rate can be applied to train JEM++ successfully, outperforming the default Adam optimizer (with a very small learning rate) used in JEM. 

\begin{table}[h!]
\caption{Ablation study of different components of JEM++. All the models are trained on CIFAR10 with $M=10$ and $N=5$.}
\label{table:ablation}\vspace{-5pt}
\begin{center}
\begin{threeparttable}
\begin{tabular}{l|c|cc}
\toprule
Ablation  & Acc \% $\uparrow$ & IS $\uparrow$ & FID $\downarrow$ \\
\midrule
JEM++              &  \bf{93.5}  &  \bf{8.29}  &  37.1   \\
w/o Proximity        &  92.9  &  7.92  &  36.0  \\
w/ Adam            &  92.5  &  7.65  &  42.7   \\
%JEM (info+BN)    & 93.0       & 8.12 & 36.8 \\
%JEM (vanilla)  & 87.1  & 7.86  & 41.5    \\
random init (w/ BN)$^1$ &  -  & -  & - \\
random init (w/o BN)$^2$  & 88.6 &  7.64  & \bf{35.1}   \\
informative init (w/o BN)$^3$ &  91.1    &  7.92  &  39.8  \\

\bottomrule
\end{tabular}
\begin{tablenotes}
  \scriptsize\item $^1$ It fails to achieve a high accuracy and generate realistic images.
  \scriptsize\item $^2$ It diverges early at epoch 28.
  \scriptsize\item $^3$ Without batch norm, only ADAM with $lr\!=\!0.0001$ can be used.
\end{tablenotes}
\end{threeparttable}
\vspace{-15px}
\end{center}
\end{table}

\vspace{-5pt}
\subsection{Classification Accuracy vs. Image Quality}

\begin{figure}[ht]
    \vspace{-5pt}
    \centering
        \includegraphics[width=0.85\columnwidth]{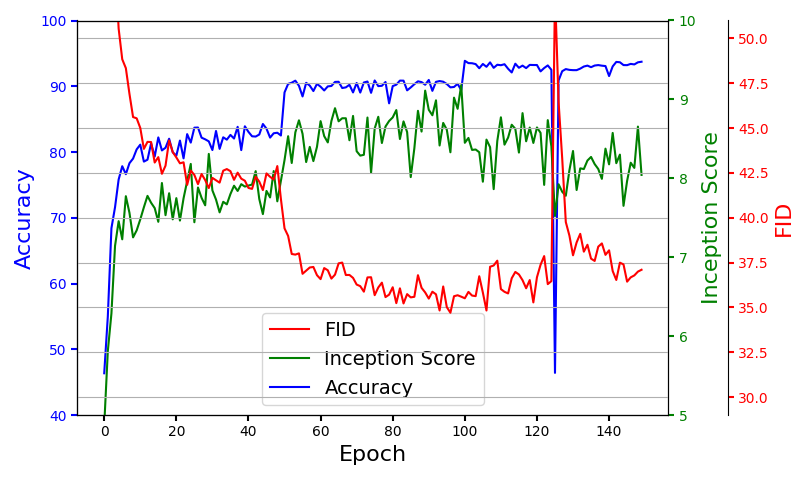}\vspace{-5pt}
    \caption{The evolution of JEM++'s classification accuracy, IS and FID scores as a function of training epochs on CIFAR10. The spike around epoch 125 is due to training instability and thanks to the proximity constraint, JEM++ stabilizes the training eventually.}
    \label{figure:acc_loss_is_fid_all}
    \vspace{-10px}
\end{figure}

One interesting phenomenon we observed from our experiments is the tradeoff between classification accuracy and image quality. Figure~\ref{figure:acc_loss_is_fid_all} shows the evolution of classification accuracy, IS and FID scores as a function of the training epochs. At the early stage of training (before epoch 100), both classification accuracy and image quality can be improved jointly. After that, there is a clear competition between accuracy and image quality, where improving accuracy hurts image quality. This probably can be explained by our minimax objective (\ref{eq:minimax}), in which the classifier and the implicit generator compete with each other to achieve an equilibrium. Compared to the standard GANs~\cite{GAN}, the difference is that we have only one network that serves both as classifier and generator. How to balance the discriminative and generative powers within one model is unclear. It would be interesting to investigate this further in the future. 

% \subsection{Best and Worst Images in Replay Buffer}

% Figure~\ref{figure:best_worst} 

% \begin{figure}[ht!]
%     \centering
%     \subfigure[Samples with highest $p(\bs{x})$]{
%         \includegraphics[width=0.45\columnwidth]{figures/cifar10_best_top10.png}
%         \label{figure:best_samples}
%     }
%     \subfigure[Samples with lowest $p(\bs{x})$]{
%         \includegraphics[width=0.45\columnwidth]{figures/cifar10_worst_top10.png}
%         \label{figure:worst_samples}
%     }
%     \caption{JEM++ generated class-conditional samples of CIFAR10. Each row corresponds to one class. }
%     \label{figure:best_worst}
% \end{figure}

%\vspace{-5pt}
\section{Conclusion}

In this paper, we propose JEM++ which improves JEM's accuracy, training stability and speed altogether with a number of new training procedures and architecture features. We demonstrate the effectiveness of these improvements on multiple benchmark datasets with state-of-the-art results in most of the tasks of image classification, image generation, adversarial robustness, uncertainty calibration and OOD detection. Most importantly, JEM++ enjoys stable and accelerated training over the original JEM.

As for future work, we plan to investigate the tradeoff between the classification accuracy and image quality as shown in Figure~\ref{figure:acc_loss_is_fid_all}. We are interested in what the optimal tradeoff is and how we can achieve the optimum with architecture design and/or new training methodologies (e.g.,~\cite{nomcmc,IBINN20}). We also plan to explore JEM++ to large-scale benchmarks, such as ImageNet, and its application to other domains, such as NLP.

\vspace{-5pt}
\section{Acknowledgment}
We would like to thank the anonymous reviewers for their comments and suggestions, which helped improve the quality of this paper. We would also gratefully acknowledge the support of VMware Inc. for its university research fund to this research.

{\small
\bibliographystyle{ieee_fullname}
\bibliography{ml}
}

\appendix
\section{Experimental Details}\label{app:exp}

To have a fair comparison with JEM~\cite{jem}, all our experiments are based on the Wide-ResNet architecture~\cite{wideresnet16} and follow JEM's settings whenever possible. As we discussed in the main text, JEM++ enables batch norm~\cite{batchnorm15} and the SGD optimizer~\cite{sgd} with a large learning rate, which we find works better than Adam~\cite{adam2014} with a very small learning rate of $1e\!-\!4$ that is used by JEM. Specifically, we use SGD with an initial learning rate of 0.1 and a decay rate of 0.2, and train all our models for 150 epochs. We reduce the learning rate at epoch [50, 100, 125].
Table~\ref{table:hyperparameters} lists the hyperparameters of JEM++. Note that JEM++ is still highly stable even with $M=5$. More experimental details can be found in our code, which is publicly available at \url{https://github.com/sndnyang/JEMPP}.

\begin{table}[ht!]
\caption{Hyperparameters of JEM++ for CIFAR10}
\label{table:hyperparameters}
\begin{center}
% \resizebox{\columnwidth}{25mm}{
\begin{threeparttable}
\begin{tabular}{l|cc}
% \hline \\[-1em]
\toprule
Variable      & Value \\
\midrule
Number of outer steps $M$            & 5, 10       \\
Number of inner steps $N$      & 5           \\
Proximity constraint $\varepsilon$  & 1           \\
Buffer size $|\mathbb{B}|$              & 10,000      \\
Reinitialization freq. $\rho$  & 5\%         \\
PYLD step-size $\alpha$         & 0.2         \\
\bottomrule
\end{tabular}
\end{threeparttable}
\end{center}\vspace{-10pt}
\end{table}

\section{Informative Initialization}

\begin{figure}[ht]
    \centering
    \subfigure[SVHN ]{
        \includegraphics[width=0.9\columnwidth]{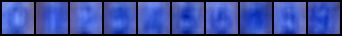}
        \label{figure:svhn_centers}
    }
    \subfigure[CIFAR100]{
        \includegraphics[width=0.9\columnwidth]{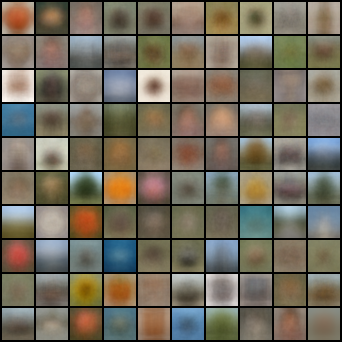}
        \label{figure:CIFAR100_centers}
    }
    \caption{The categorical centers of SVHN and CIFAR100.}
    \label{figure:informative_centers}
\end{figure}

\begin{figure}[ht]
    \centering
    \subfigure[Categorical centers of CIFAR10]{
        \includegraphics[width=0.9\columnwidth]{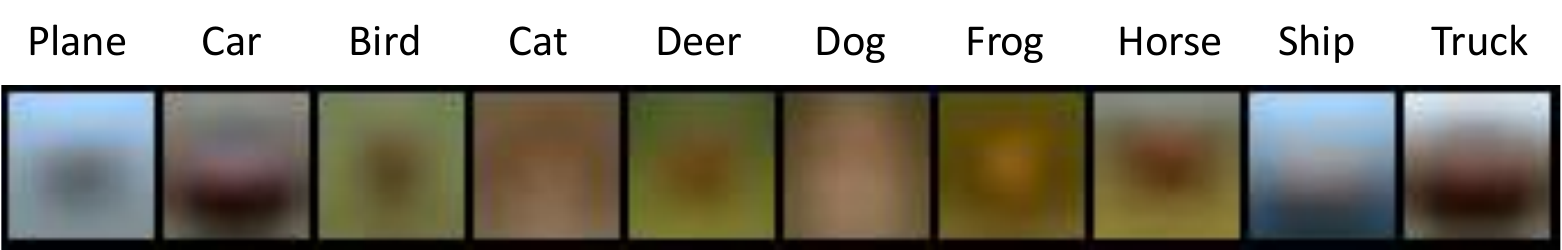}
        \label{figure:CIFAR10_centers}
    }
    \subfigure[Samples from each category]{
        \includegraphics[width=0.9\columnwidth]{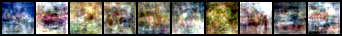}
        \label{figure:CIFAR10_init_examples}
    }
    \caption{The categorical centers of and corresponding samples of CIFAR10.}
    \label{figure:cifar10_centers_samples}\vspace{-5pt}
\end{figure}

In this paper, we introduce a novel informative initialization to start the SGLD chain. Specifically, instead of using a uniform distribution, we sample from a Gaussian mixture
distribution estimated from the training data as
\begin{align}\label{eq:gmm}
  p_0(\bs{x})&=\sum\nolimits_y\mathcal\pi_y{N}(\bs{\mu}_y,\bs{\Sigma}_y)\\
  \text{with}\quad \pi_y &= |\mathcal{D}_y|/\sum\nolimits_{y'}|\mathcal{D}_{y'}|,\quad
\bs{\mu}_{y} = \mathbb{E}_{\bs{x} \sim \mathcal{D}_{y}}[\bs{x}], \nonumber \\ 
\bs{\Sigma}_y&=\mathbb{E}_{\bs{x} \sim \mathcal{D}_{y}}\left[\left( \bs{x} - \bs{\mu}_{y}\right)\left( \bs{x} -\bs{\mu}_{y}\right)^{\top}\right],\nonumber
\end{align}
where $\mathcal{D}_{y}$ denotes the set of training samples with label $y$. Figure~\ref{figure:informative_centers} visualizes the categorical centers ($\bs{\mu}$'s) estimated from the SVHN and CIFAR100 training datasets.  Figure~\ref{figure:cifar10_centers_samples} visualizes the categorical centers and the corresponding samples $\bs{x}_0\sim p_0(\bs{x})$ for CIFAR10. Note that no extra information is used to train JEM++ over JEM.

% \section{Convergence Rate}

% We investigate the convergence rate of JEM and JEM++. To make a fair comparison, we train JEM++ using Adam with $lr=0.0001$, the same as JEM. Figure~\ref{figure:converge_curve} demonstrates that JEM++ converges faster than JEM.

% \vspace{-10pt}
% \begin{figure}[ht]
%     \centering
%     \subfigure[Accuracy]{
%         \includegraphics[width=0.45\columnwidth]{figures/acc_converge_jempp_jem.png}
%         \label{figure:acc_converge}
%     }
%     \subfigure[Loss]{
%         \includegraphics[width=0.45\columnwidth]{figures/loss_converge_jempp_jem.png}
%         \label{figure:loss_converge}
%     }
%     \caption{The convergence comparison between JEM and JEM++.}
%     \label{figure:converge_curve}
%     \vspace{-10pt}
% \end{figure}

\section{Applications}
In the main text, we compared JEM and JEM++ in terms of classification accuacy, image quality, training stability and speed. Here we compare JEM and JEM++ in other downstream applications, such as adversarial robustness, calibration and out of distribution (OOD) detection.

\vspace{-0pt}
\subsection{Robustness}

It's well known that DNNs are particularly vulnerable to adversarial examples~\cite{propertiesNN14, goodfellow2014explaining} in the form of small perturbations to inputs that lead DNNs to predict incorrect outputs.  %Szegedy et al.~\cite{propertiesNN14} found that very tiny perturbations to input samples (a.k.a., adversarial examples) can easily fool a well-trained DNN. 
Specially, the widely explored adversarial examples are defined as perturbed inputs $\tilde{\bs{x}} = \bs{x} + \bs{\delta}$ under an $L_p$-norm constraint $\| \bs{\delta}\|_p<\varepsilon$. %The models are easily to make mistakes on adversarial examples when asked to classify inputs it has rarely or never encountered during training. 
To overcome the security threat posed by adversarial examples, a variety of defense algorithms have been proposed in the past few years to improve the robustness of models~\cite{advexample15,dziugaite2016study,guo2017countering,akhtar2018defense,madry2018towards, chiang2020certified}. Among them, adversarial training~\cite{advexample15,madry2018towards} has been proved to be the most effective one to defend adversarial examples. 

As we discussed in Section 3.1, there is a close relationship between the maximum likelihood learning of EBM (7) and adversarial training with PGD~\cite{madry2018towards} as both solve a similar minimax objective. Therefore, the maximum likelihood trained EBMs should be more robust to adversarial examples than the standard trained softmax classifiers, and this has been empirically verified by recent works (e.g.,~\cite{du2019implicit,jem}). Since JEM++ improves JEM's accuracy, training stability and speed, it's interesting to check if JEM++ can improve model robustness as well.

%What's more, another type of methods is to detect and robustly classify adversarial examples with generative models~\cite{song2017pixeldefend,schott2018towards}. This approach makes the model exposed to sufficient ``unseen" data and thus the model is able to accurately classify and defense adversarial examples. Given the facts that an EBM embeds the capacity of a generative model into the classifier and the SGLD training of EBMs shows quite similarity to adversarial training, 

\vspace{-10pt}
\begin{figure}[ht]
    \centering
    \subfigure[$L_\infty$ Robustness]{
        \includegraphics[width=0.45\columnwidth]{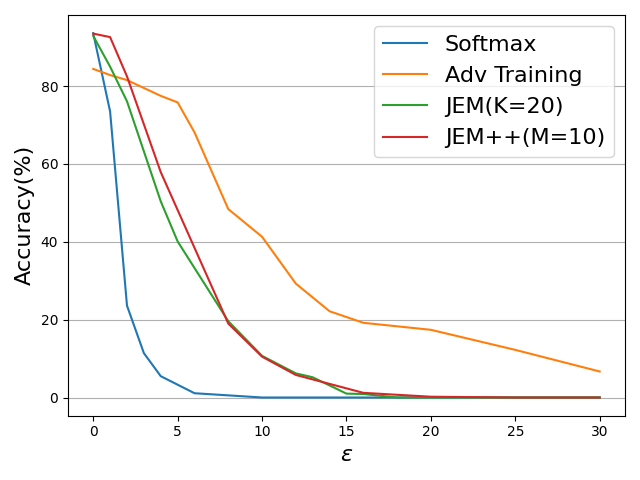}
        \label{figure:l_inf_robust}
    }
    \subfigure[$L_2$ Robustness]{
        \includegraphics[width=0.45\columnwidth]{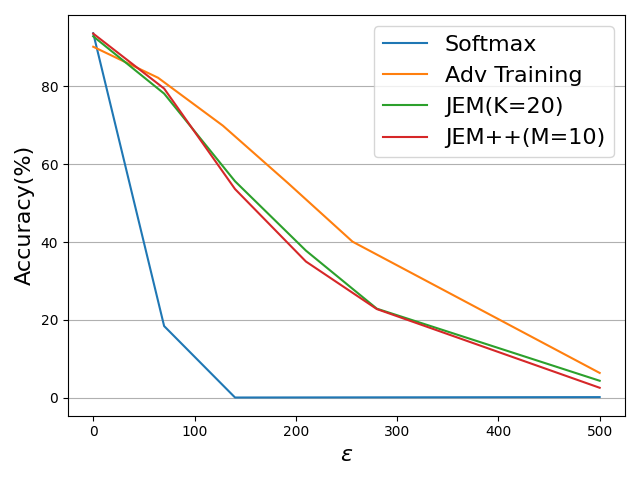}
        \label{figure:l2_robust}
    }
    \caption{Adversarial robustness under the PGD attacks.}
    \label{figure:robust_curve}
    \vspace{-10pt}
\end{figure}

To evaluate the robustness of a given model, we run a white-box PGD attack~\cite{madry2018towards} under an $L_\infty$ or $L_2$ constraint using foolbox~\cite{rauber2017foolbox}, with the results reported in Figure~\ref{figure:robust_curve}. It can be observed that JEM++ achieves a similar robustness with JEM under the $L_\infty$ and $L_2$ PGD attacks, while both are more robust than the standard softmax classifiers. The adversarial training with PGD~\cite{madry2018towards,santurkar2019image} achieves the highest robustness since it is trained and test under the same PGD attacks, while JEM/JEM++ are trained on real and generated samples from the energy function, without the access to the PGD samples for training.

\subsection{Calibration}
Recent researches have shown that the predictions from modern DNNs could be over-confident~\cite{guo2017calibration}, i.e., they often output incorrect but confident predictions, which could have catastrophic consequences. Hence, calibration of uncertainty for DNNs is a critical task with an enormous practical  impact nowadays. Here, the confidence is defined as $\max_y p(y|\bs{x})$ which is used to decide when to output a prediction. In this section, we compare the calibration qualities of models trained by JEM and JEM++ as well as the standard softmax classifiers on the CIFAR10/100 dataset.

\vspace{-5pt}
\paragraph{Expected Calibration Error} (ECE) is a standard metric to evaluate the calibration quality of a classifier~\cite{guo2017calibration}. It firstly computes the confidence of the model, $\max_y p(y|\bs{x}_i)$, for each $\bs{x}_i$ in the dataset. Then it groups the predictions into equally spaced buckets $\{B_1,B_2,\cdots, B_M\}$ based on the confidence scores. For example, if $M$ = 20, then $B_1$ would represent all examples for which the model's confidence scores were between 0 and 0.05. Then ECE is calculated as\vspace{-5pt}
\begin{equation}
    \mathrm{ECE}=\sum_{m=1}^{M} \frac{\left|B_{m}\right|}{n}\left|\operatorname{acc}\left(B_{m}\right)-\operatorname{conf}\left(B_{m}\right)\right|,
\end{equation}
where $n$ is the number of data in the dataset, acc($B_m$) is the average accuracy of the model on all the examples in $B_m$ and conf($B_m$) is the average confidence on all the examples in $B_m$. In our experiments, we set $M$ = 20. For a perfectly calibrated model, the ECE will be 0 for any $M$.

\begin{figure}[ht!]
    \centering
    \subfigure[Standard Softmax]{
        \includegraphics[width=0.45\columnwidth]{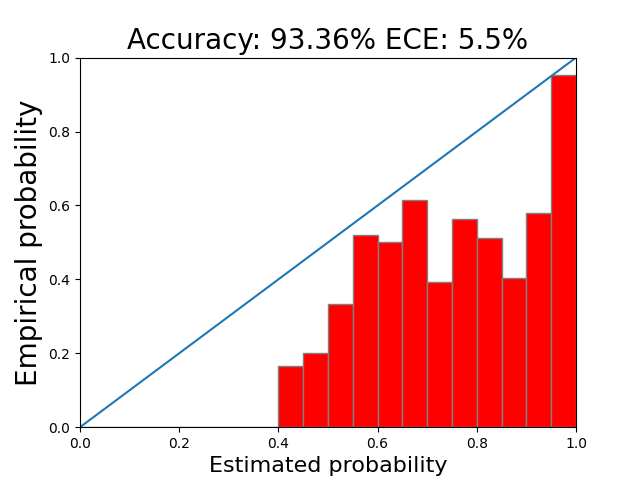}
        \label{figure:base_cali}
    }
    \subfigure[JEM (K=20)]{
        \includegraphics[width=0.45\columnwidth]{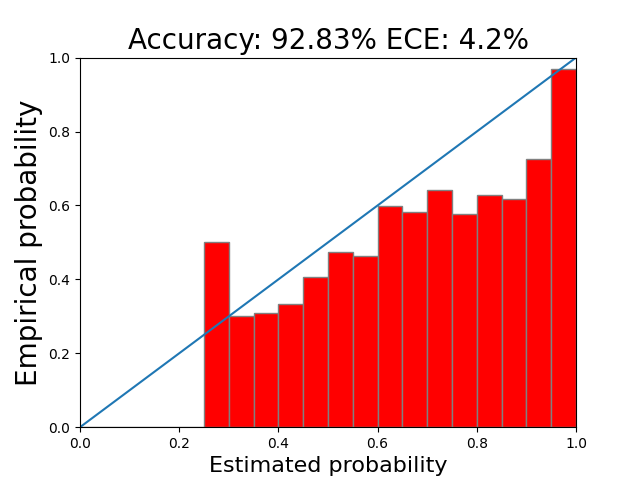}
        \label{figure:jem_cali}
    }
    \subfigure[JEM++ (M=5)]{
        \includegraphics[width=0.45\columnwidth]{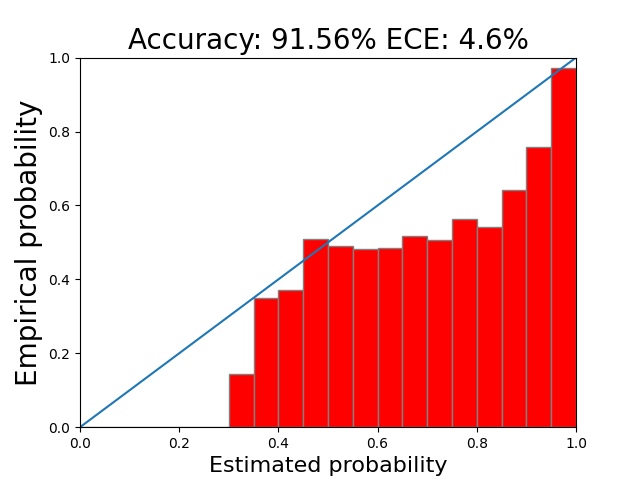}
        \label{figure:jempp_M5_cali}
    }
    \subfigure[JEM++ (M=10)]{
        \includegraphics[width=0.45\columnwidth]{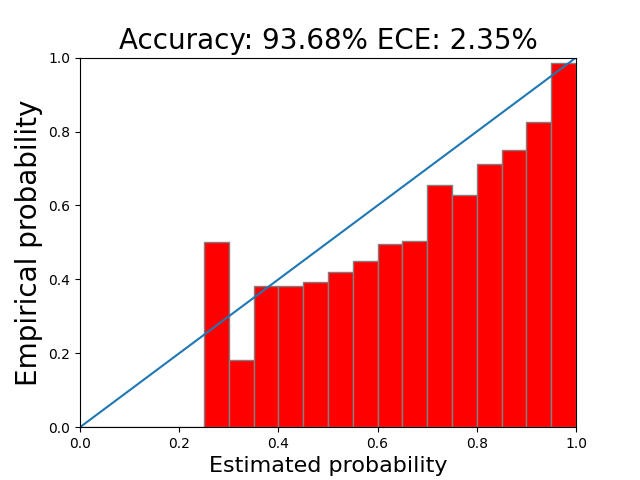}
        \label{figure:jempp_M10_cali}
    }
    \caption{Calibration results on CIFAR10. The smaller ECE is, the better.}
    \label{figure:CIFAR10_cali}
\end{figure}

\begin{figure}[ht!]
    \centering
    \subfigure[Standard Softmax]{
        \includegraphics[width=0.45\columnwidth]{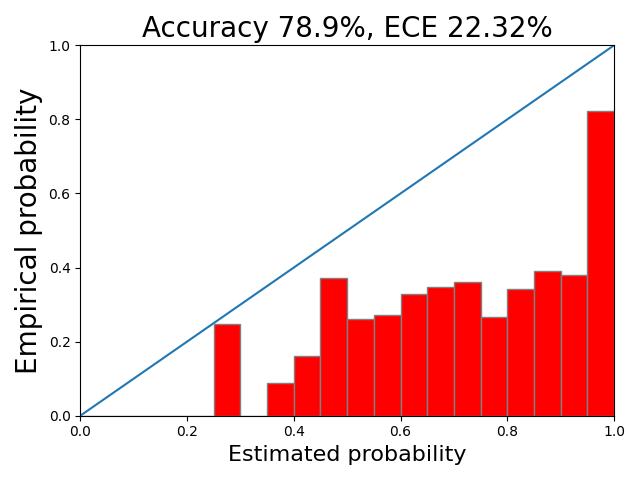}
        \label{figure:base_CIFAR100_cali}
    }
    \subfigure[JEM (K=20)]{
        \includegraphics[width=0.45\columnwidth]{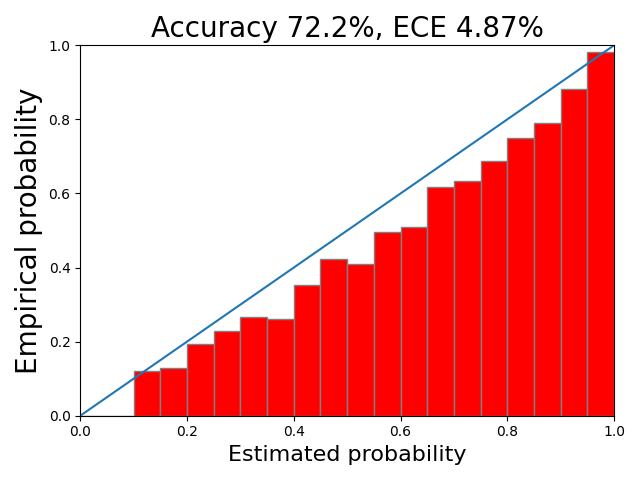}
        \label{figure:jem_CIFAR100_cali}
    }
    \subfigure[JEM++ (M=5)]{
        \includegraphics[width=0.45\columnwidth]{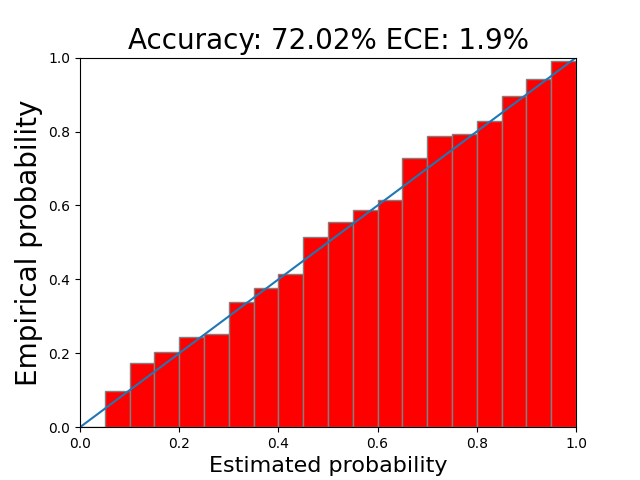}
        \label{figure:jempp_CIFAR100_m5_cali}
    }
    \subfigure[JEM++ (M=10)]{
        \includegraphics[width=0.45\columnwidth]{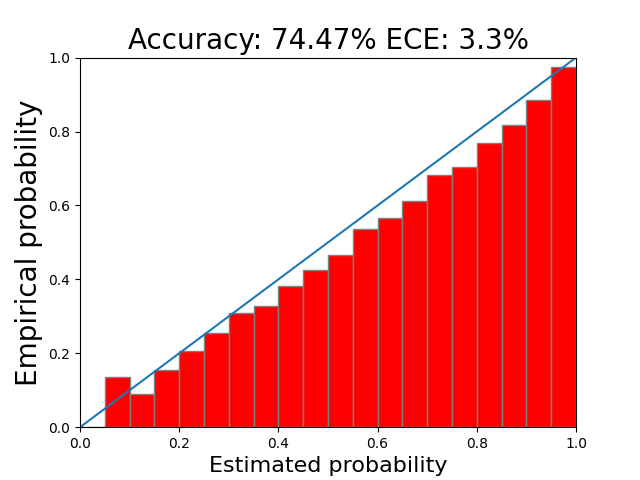}
        \label{figure:jempp_CIFAR100_m10_cali}
    }
    \caption{Calibration results on CIFAR100. The smaller ECE is, the better.}
    \label{figure:CIFAR100_cali}
\end{figure}

Figures~\ref{figure:CIFAR10_cali} and~\ref{figure:CIFAR100_cali} report the results on CIFAR10 and CIFAR100, respectively. As we can see, the models trained by JEM and JEM++ are better calibrated than the standard softmax classifiers, while JEM++ achieves better calibration qualities than JEM on CIFAR10 (2.35\% vs. 4.2\%) and CIFAR100 (3.3\% vs. 4.87\%) with notable margins.

\subsection{Out-Of-Distribution Detection}

The OOD detection is a binary classification problem, which outputs a score $s_{\bs{\theta}}(\bs{x}) \in \mathbb{R}$ for a given query $\bs{x}$. The model should be able to assign lower scores to OOD examples than to in-distribution examples, such that it can be used to distinguish two sets of examples. Following the settings of JEM~\cite{jem}, we use the Area Under the Receiver-Operating Curve (AUROC)~\cite{HenGim16} to evaluate the performance of OOD detection. In our experiments, two standard score functions are considered: the input density $p_{\bs{\theta}}(\bs{x})$~\cite{nalisnick2018deep} and the predictive distribution $p_{\bs{\theta}}(y|\bs{x})$~\cite{HenGim16}.

\paragraph{Input Density}

A natural choice of $s_{\bs{\theta}}(\bs{x})$ is the input density $p_{\bs{\theta}}(\bs{x})$. For OOD detection, intuitively we consider examples with low $p(\bs{x})$ to be OOD. Quantitative results can be found in Table~\ref{CIFAR10_ood} (top row), where CIFAR10 is the in-distribution data and SVHN, an interpolated CIFAR10, CIFAR100 and CelebA are treated as out-of-distribution data, respectively. Moreover, the corresponding distributions of scores are visualized in Table~\ref{table:logpx_hist}. As can be seen, the JEM++ model assigns higher likelihoods to in-distribution data than to the OOD data, outperforming JEM and all the other models by significant margins.

\paragraph{Predictive Distribution}

Another useful OOD score is the maximum probability from a classifier's predictive distribution: $s_{\bs{\theta}}(\bs{x}) = \max_y p_{\bs{\theta}}(y|\bs{x})$. Hence, OOD performance using this score is highly correlated with a model's classification accuracy. The results can be found in Table~\ref{CIFAR10_ood} (bottom row). Again, JEM++ outperforms JEM and all the other models by notable margins.

%In summary, among two different OOD score functions, JEM++ outperforms all the competing methods as can be seen in Table~\ref{CIFAR10_ood} while different $M$ show a tradeoff of performance.

\begin{table*}[ht!]
\caption{OOD detection results. Models are trained on CIFAR10. Values are AUROC.}
\label{CIFAR10_ood}
\begin{center}
\begin{threeparttable}
\begin{tabular}{c|c|cccc}
\toprule
$s_{\bs{\theta}}(\bs{x})$  & Model   & SVHN & CIFAR10 Interp & CIFAR100 & CelebA \\
\midrule
\multirow{5}{*}{$\log p_{\bs{\theta}}(\bs{x})$} & Uncond Glow          & .05 & .51 & .55 & .57 \\
            %& Class-Conditional Glow      & .07 & .45 & .51 & .53 \\
            & IGEBM                  & .63 & .70 & .50 & .70 \\
            & JEM (K=20)             & .67 & .65 & .67 & .75 \\
            & JEM++ (M=5)            & \bf{.89} & \bf{.73} & \bf{.81} & .74 \\
            & JEM++ (M=10)           & .63 & .68 & .64 & .59  \\
            & JEM++ (M=20)           & .85 & .57 & .68 & \bf{.89} \\
\midrule
\multirow{5}{*}{$\max_y p_{\bs{\theta}}(y|\bs{x})$} & WideResNet             & .93 & .77 & .85 & .62 \\
    %& Class-Conditional Glow  & .64 & .61 & .65 & .54 \\
    & IGEBM                   & .43 & .69 & .54 & .69 \\
    & JEM (K=20)              & .89 & .75 & .87 & .79 \\
    & JEM++ (M=5)             & .88 & .78 & .86 & .78 \\
    & JEM++ (M=10)            & .91 & \bf{.78} & .88 & .82    \\
    & JEM++ (M=20)            & \bf{.94} & .77 & \bf{.88} & \bf{.90} \\
\bottomrule
\end{tabular}
\end{threeparttable}
\end{center}
\end{table*}

\begin{table*}[ht!]
  \centering
  \begin{tabular}{ | c | m{4.5cm} | m{4.5cm} | m{4.5cm} | }
    \hline
    JEM
    &
    \begin{minipage}{.27\textwidth}
      \includegraphics[width=\linewidth, height=32mm]{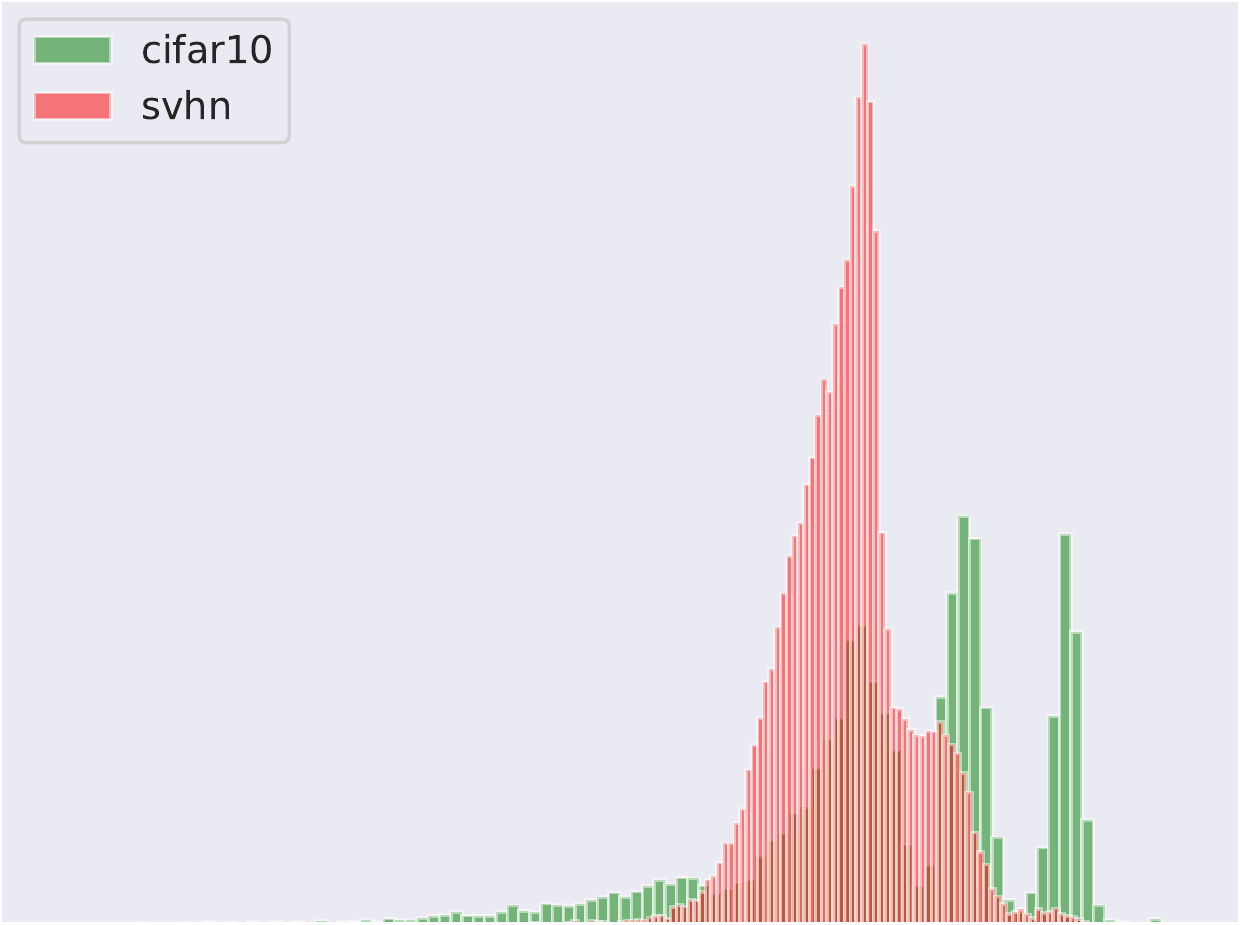}
    \end{minipage}
    &
    \begin{minipage}{.27\textwidth}
      \includegraphics[width=\linewidth, height=32mm]{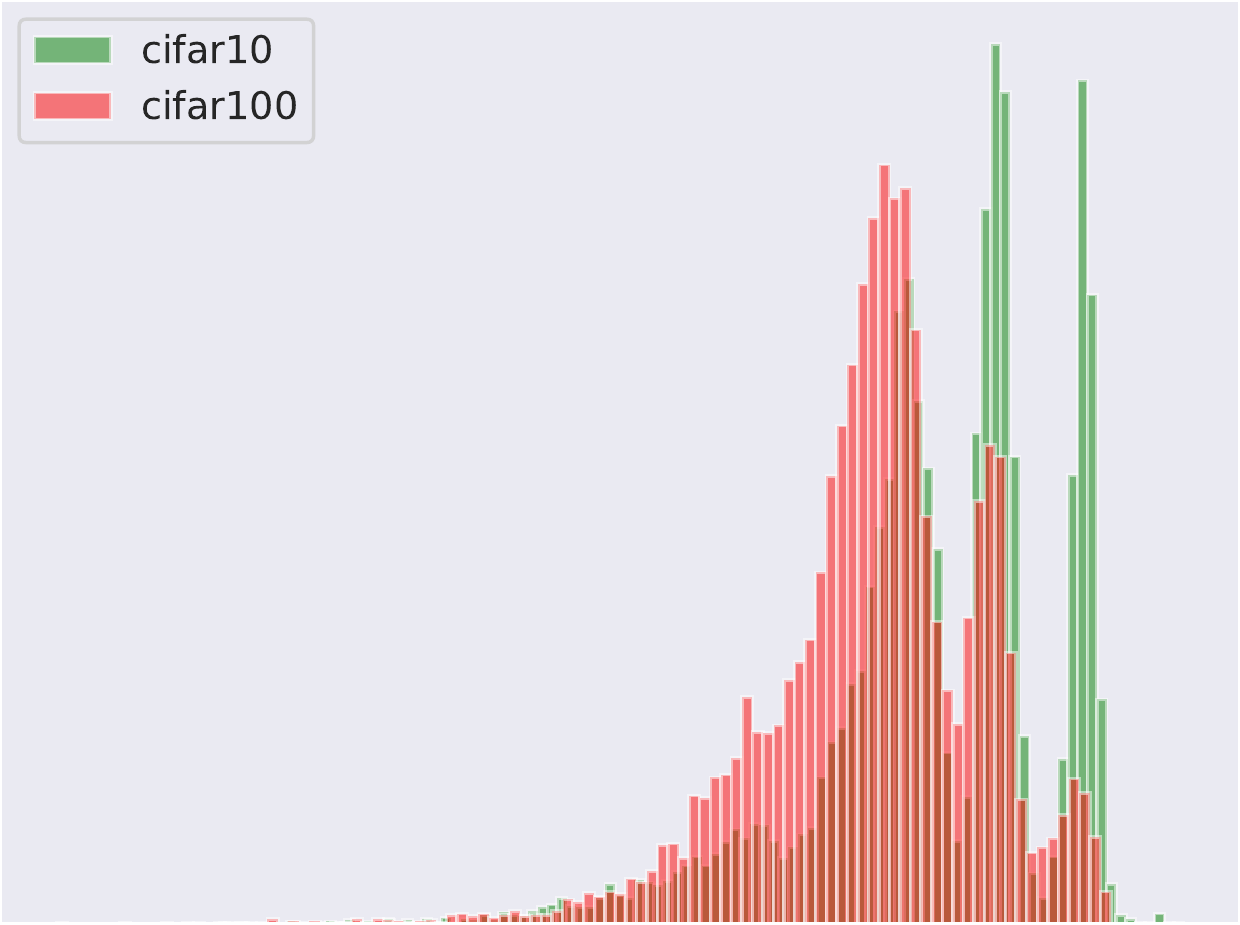}
    \end{minipage}
    &
    \begin{minipage}{.27\textwidth}
      \includegraphics[width=\linewidth, height=32mm]{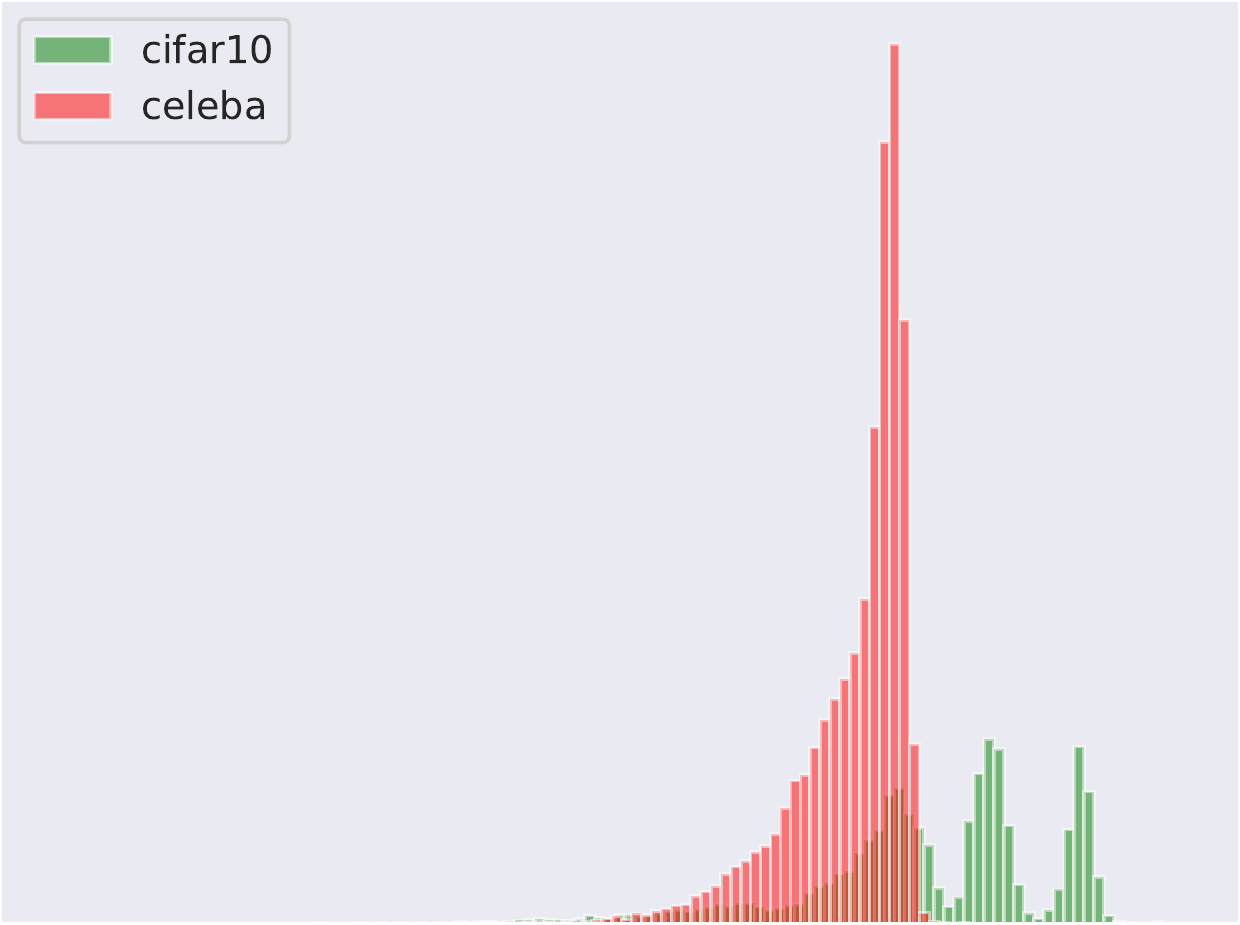}
    \end{minipage}
      \\ \hline
    JEM++(M=5)
    &
    \begin{minipage}{.27\textwidth}
      \includegraphics[width=\linewidth, height=32mm]{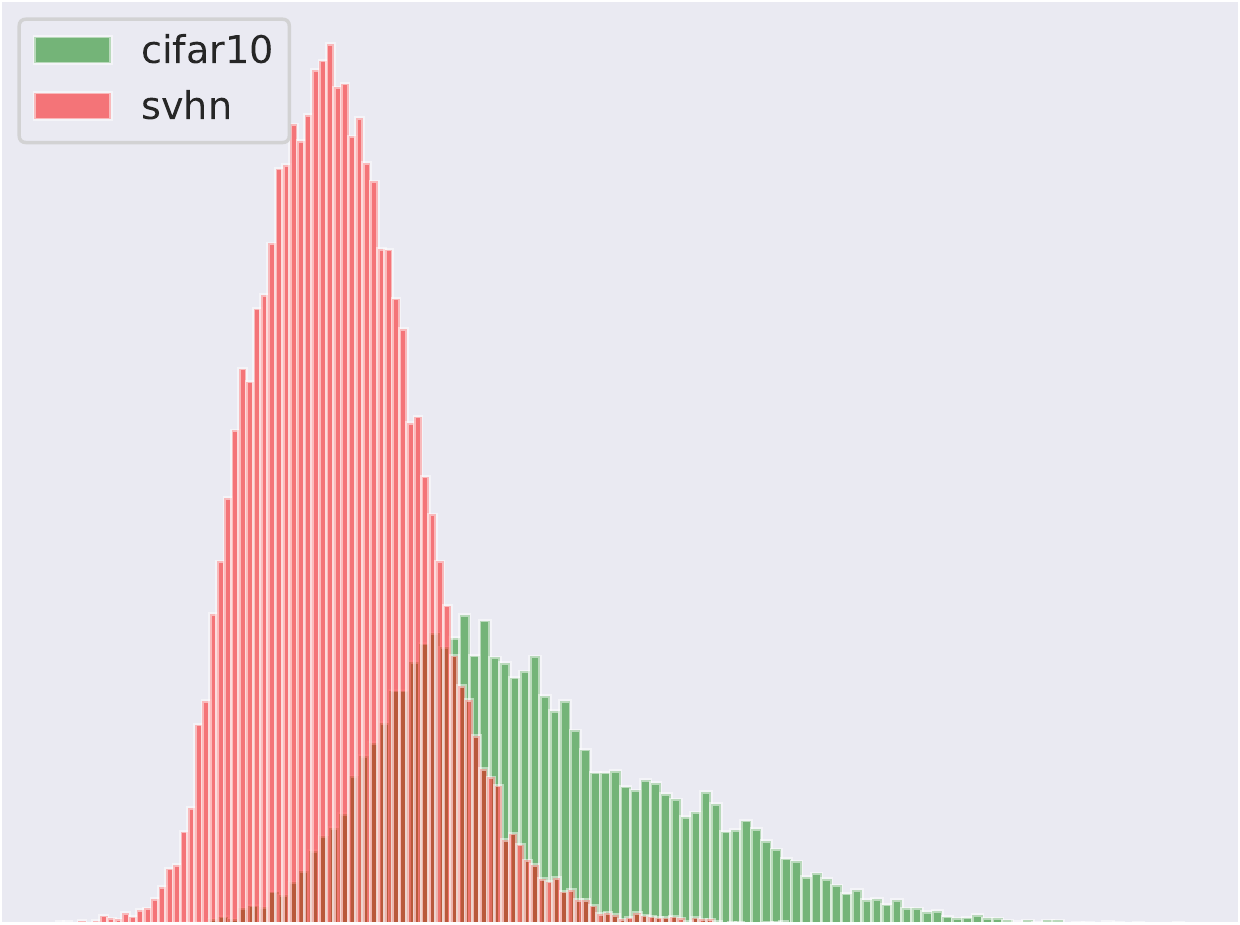}
    \end{minipage}
    &
    \begin{minipage}{.27\textwidth}
      \includegraphics[width=\linewidth, height=32mm]{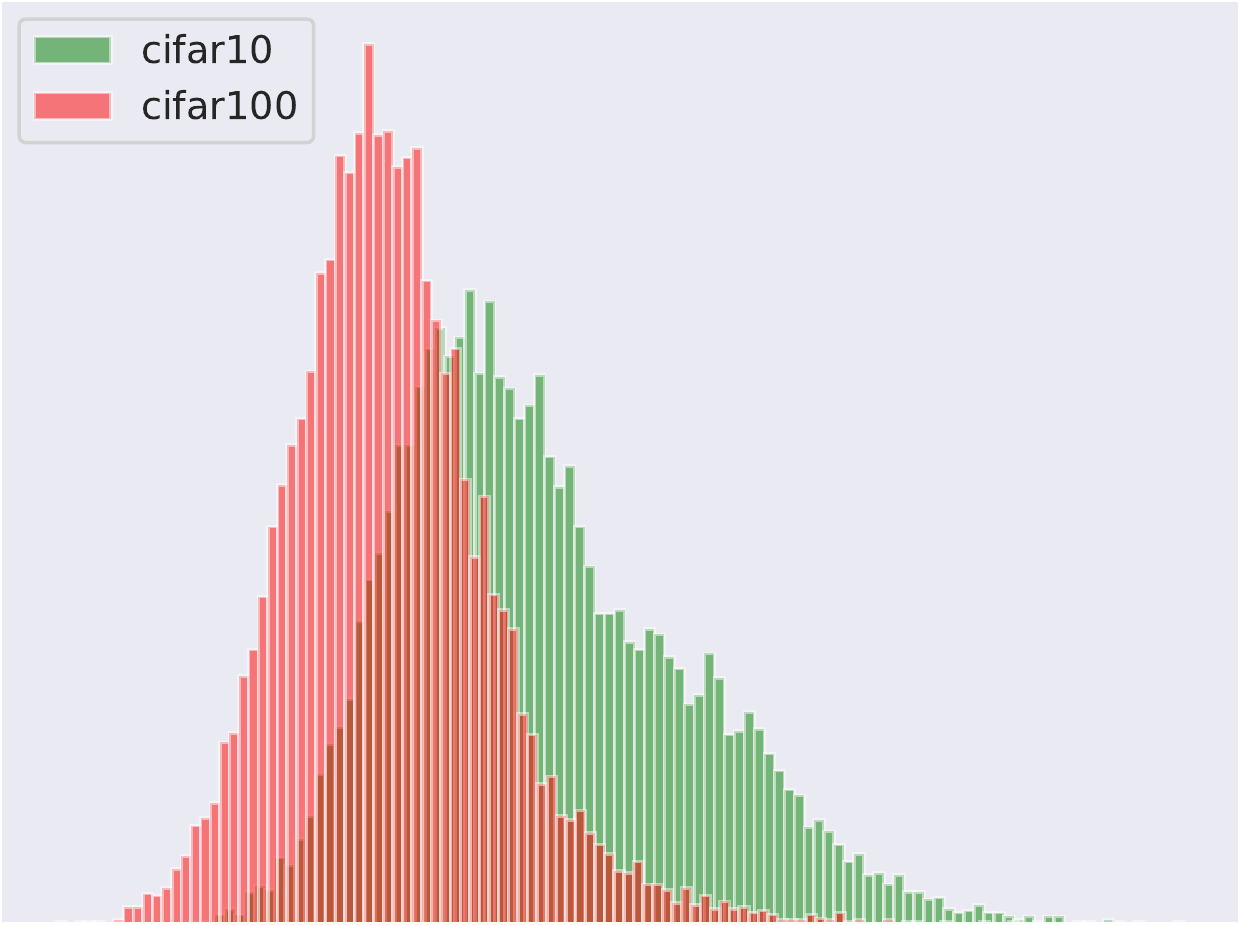}
    \end{minipage}
    &
    \begin{minipage}{.27\textwidth}
      \includegraphics[width=\linewidth, height=32mm]{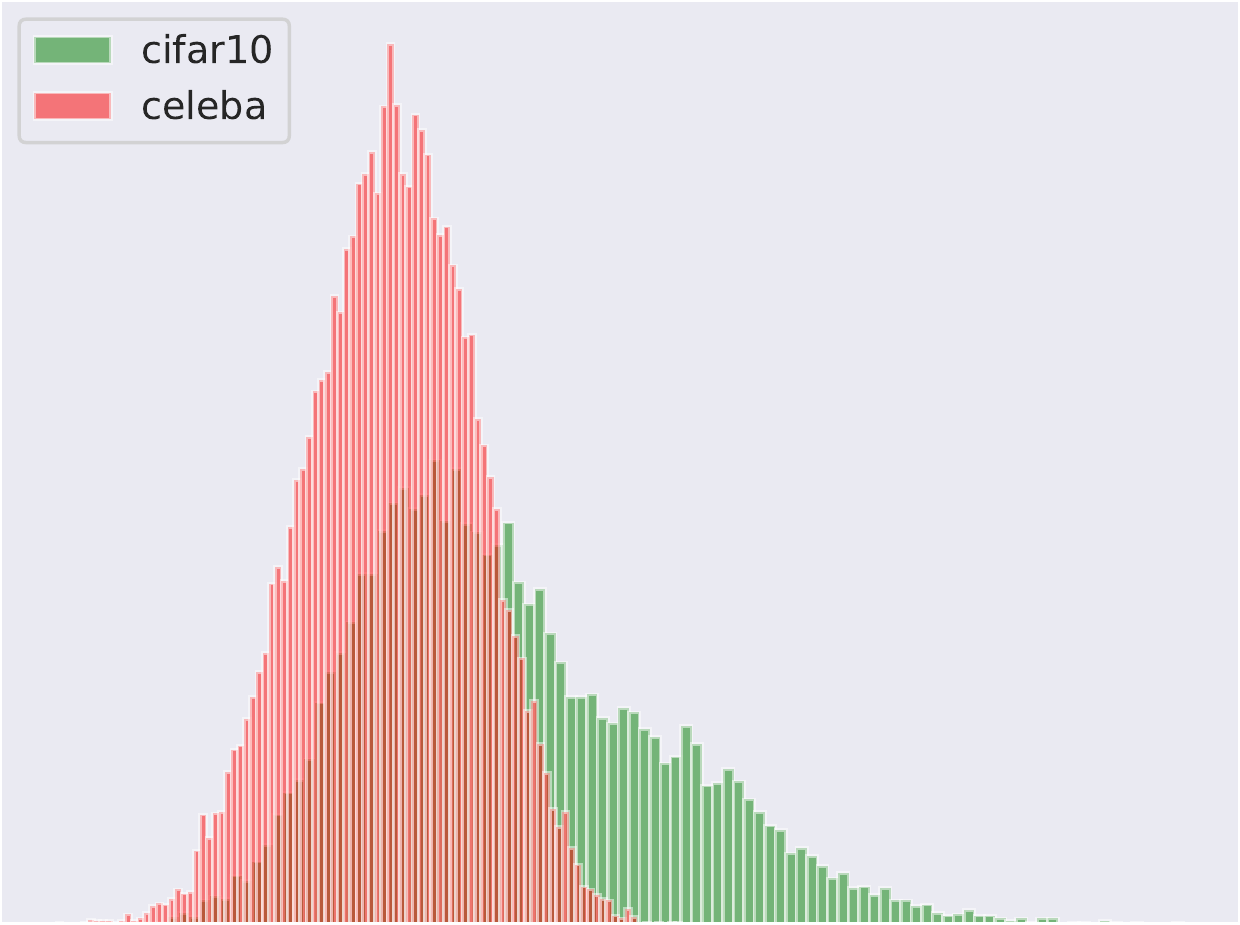}
    \end{minipage}
      \\ \hline
    JEM++(M=10)
    &
    \begin{minipage}{.27\textwidth}
      \includegraphics[width=\linewidth, height=38mm]{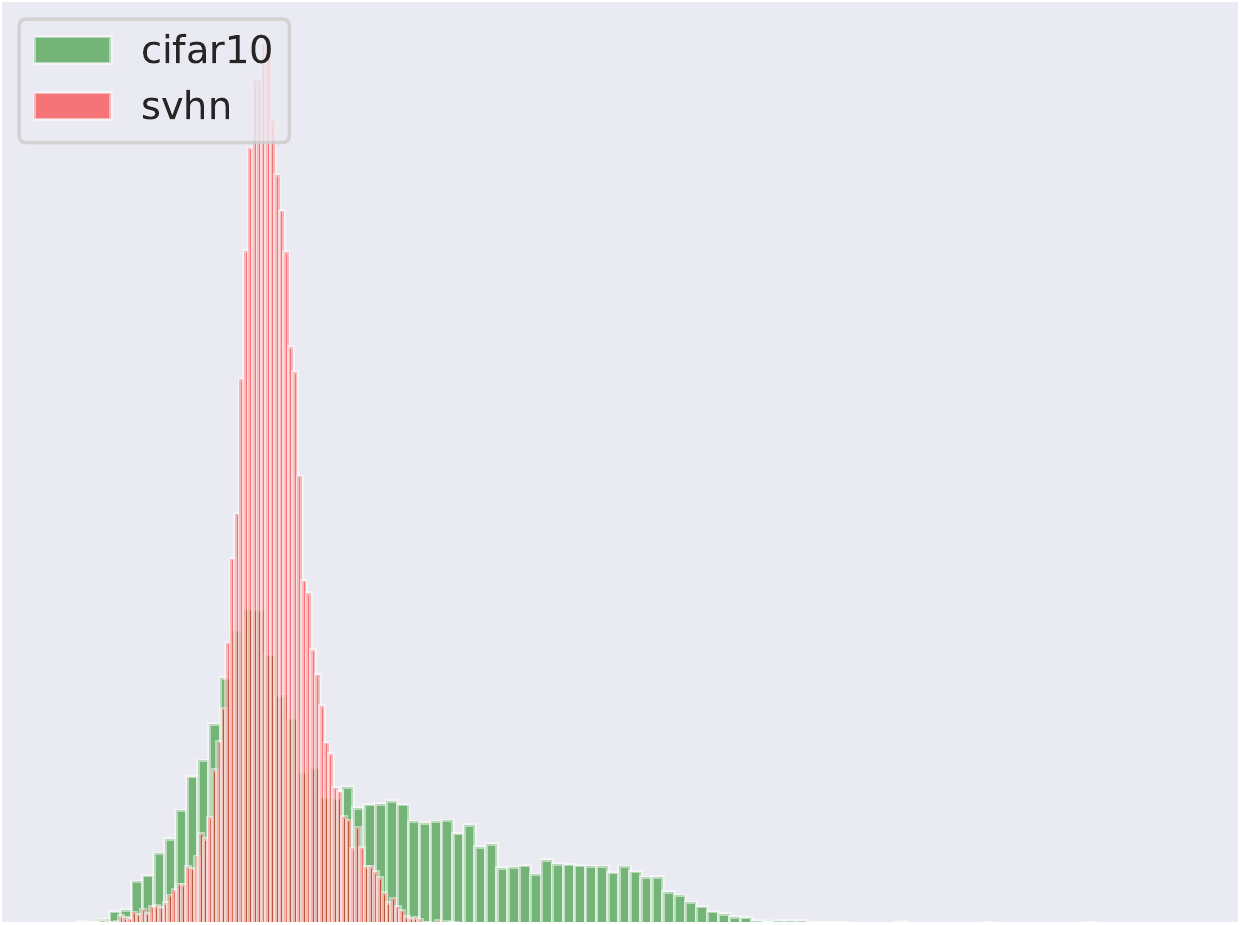}
    \end{minipage}
    &
    \begin{minipage}{.27\textwidth}
      \includegraphics[width=\linewidth, height=38mm]{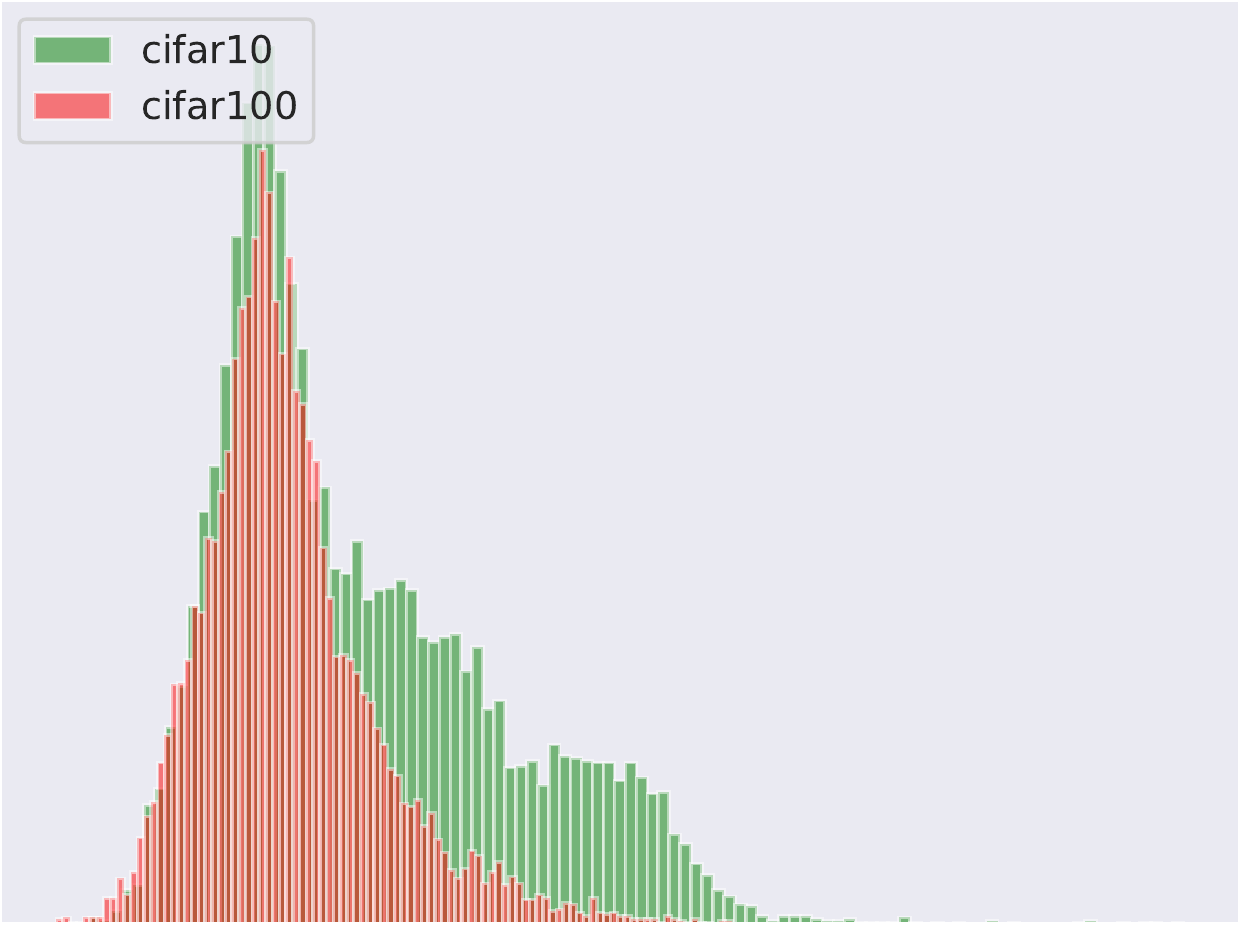}
    \end{minipage}
    &
    \begin{minipage}{.27\textwidth}
      \includegraphics[width=\linewidth, height=38mm]{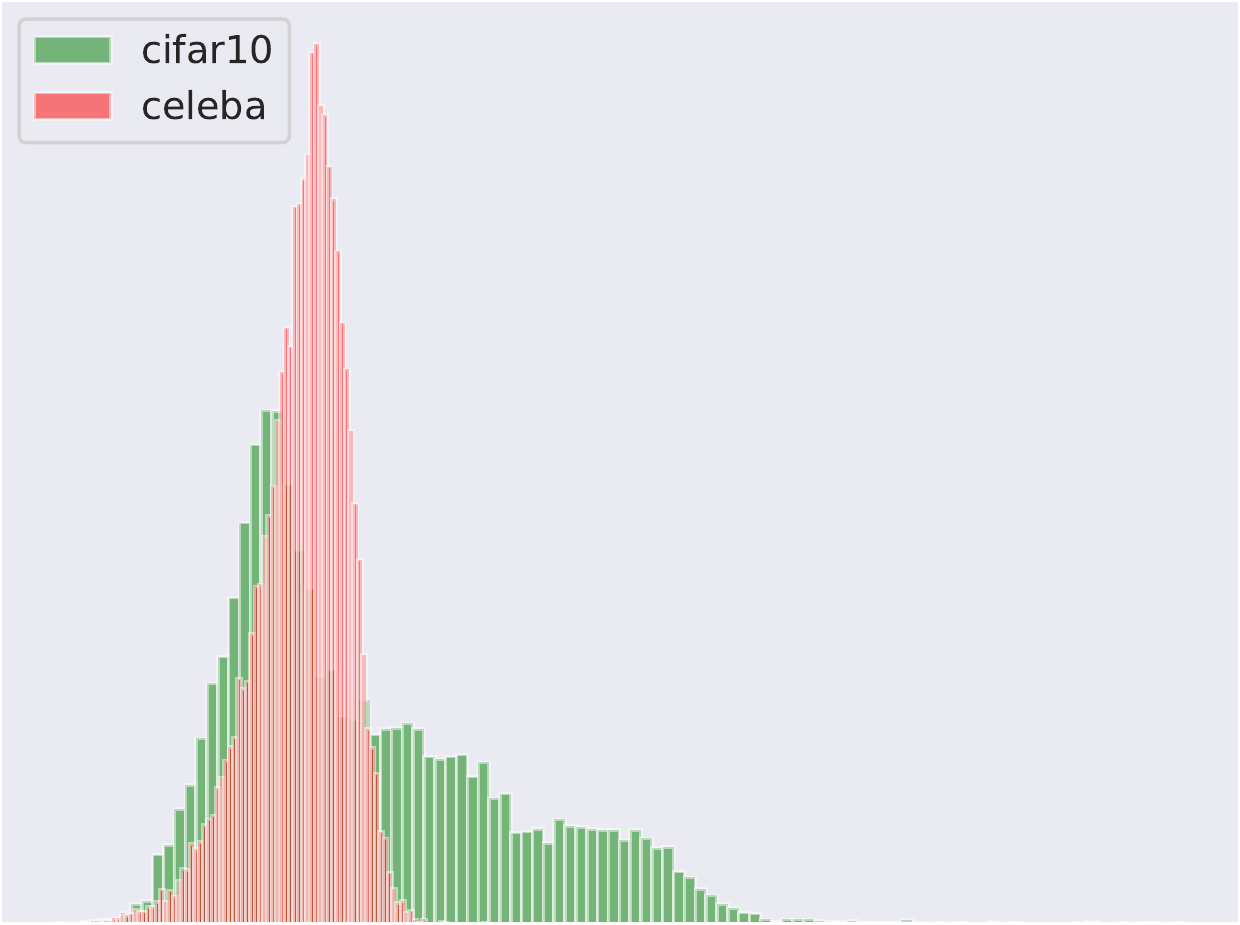}
    \end{minipage}
      \\ \hline
     JEM++(M=20)
    &
    \begin{minipage}{.27\textwidth}
      \includegraphics[width=\linewidth, height=38mm]{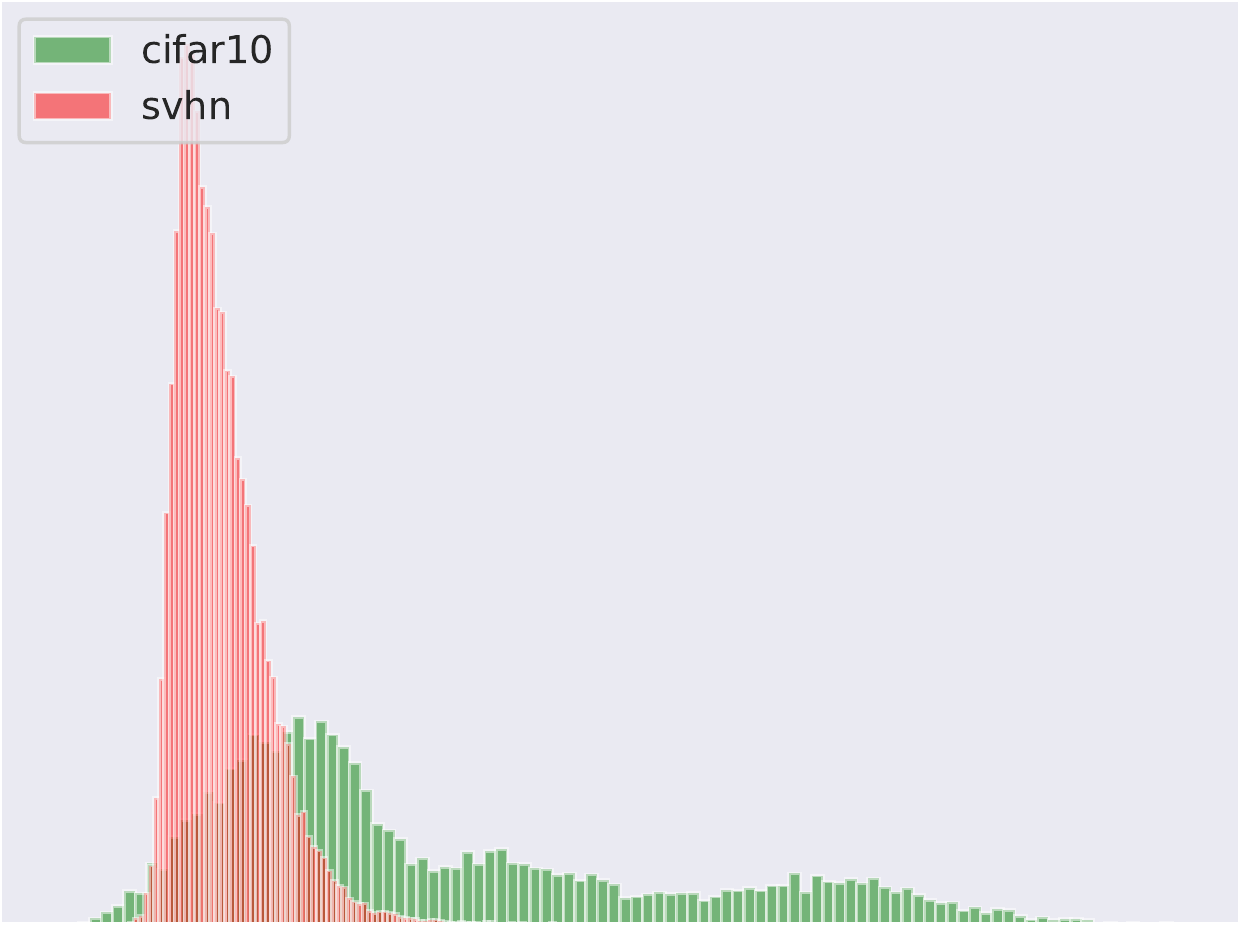}
    \end{minipage}
    &
    \begin{minipage}{.27\textwidth}
      \includegraphics[width=\linewidth, height=38mm]{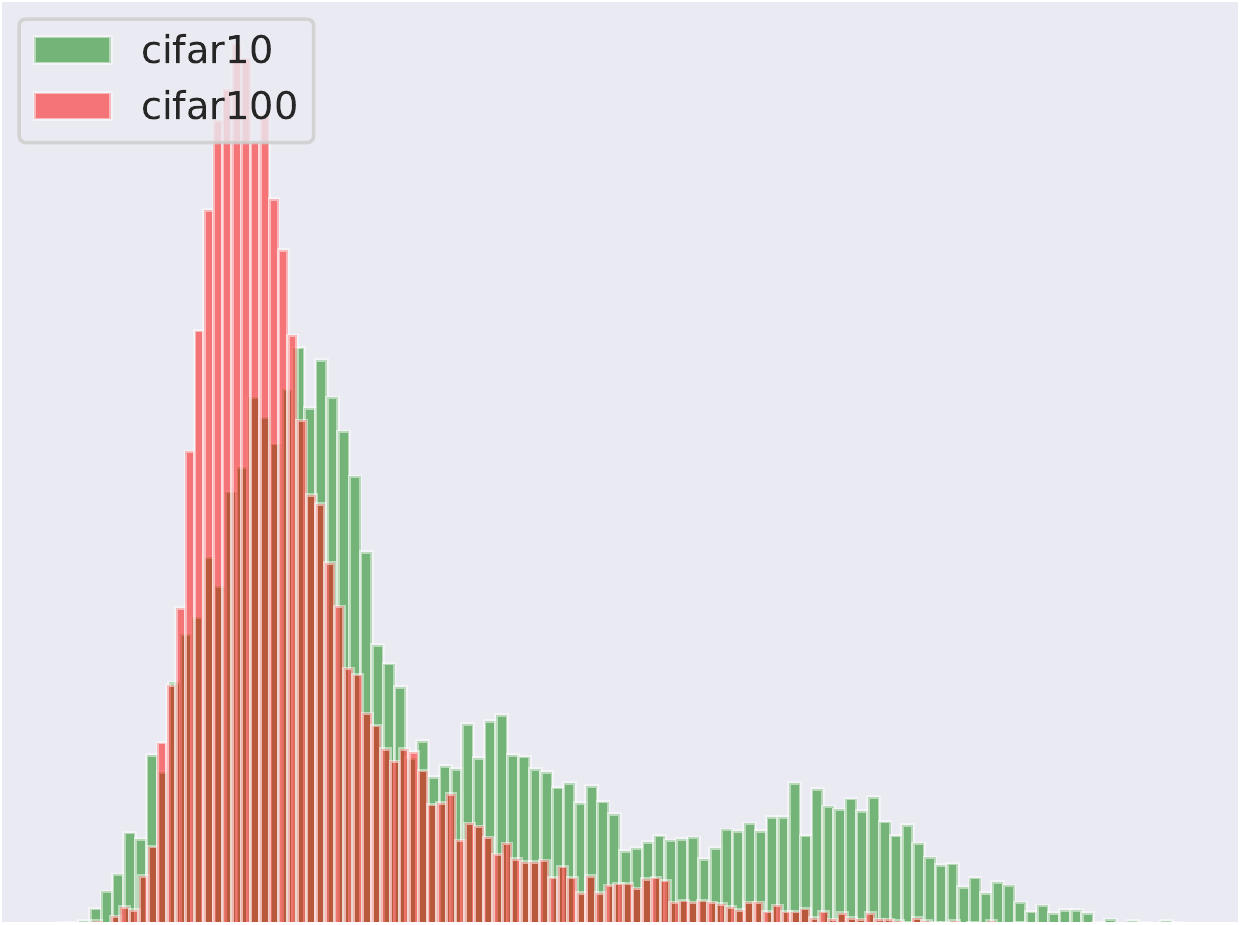}
    \end{minipage}
    &
    \begin{minipage}{.27\textwidth}
      \includegraphics[width=\linewidth, height=38mm]{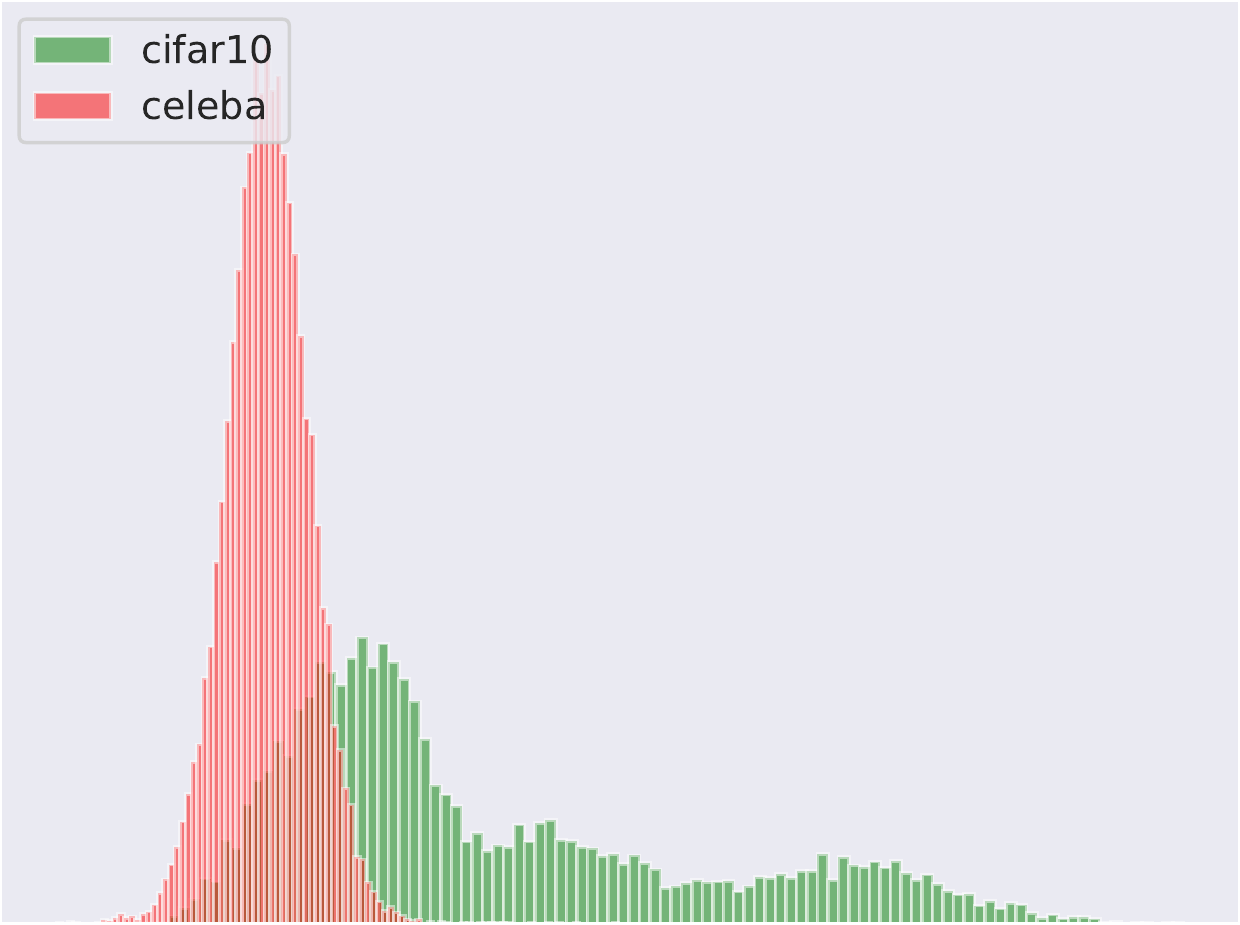}
    \end{minipage}
      \\ \hline
  \end{tabular}
  \caption{Histograms of $\log_{\bs{\theta}}p(\bs{x})$ for OOD detection. Green corresponds to in-distribution dataset, while red corresponds to OOD dataset.} %All models trained on CIFAR10. }
  \label{table:logpx_hist}
\end{table*}

\section{Additional Generated Samples}\label{app:samples}

\begin{figure}[ht!]
    \centering
    \subfigure[SVHN (Conditional)]{
        \includegraphics[width=0.45\columnwidth]{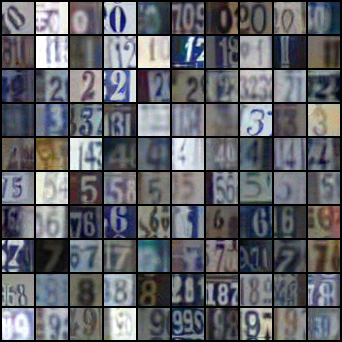}
        \label{figure:svhn_cond}
    }
    \subfigure[CIFAR100 (Conditional)]{
        \includegraphics[width=0.45\columnwidth]{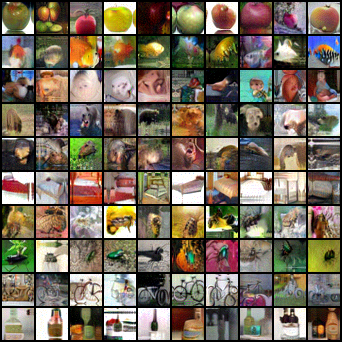}
        \label{figure:CIFAR100_cond}
    }
    \caption{JEM++ generated class-conditional samples of SVHN and CIFAR100. Each row corresponds to one class. }
    \label{figure:svhn_CIFAR100_cond}
\end{figure}

\begin{figure*}[ht!]
    \centering
    \subfigure[Samples with highest $p(\bs{x}$)]{
        \includegraphics[width=0.48\columnwidth]{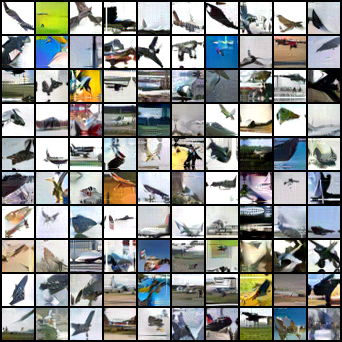}
    }
    \subfigure[Samples with lowest $p(\bs{x})$]{
        \includegraphics[width=0.48\columnwidth]{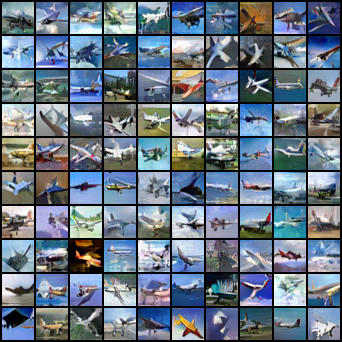}
    }
    \subfigure[Samples with highest $p(y|\bs{x}$)]{
        \includegraphics[width=0.48\columnwidth]{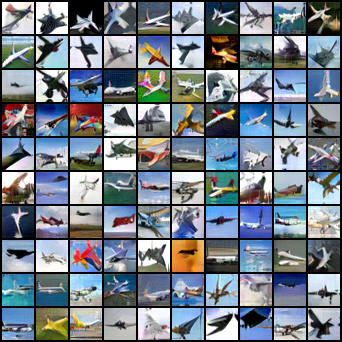}
    }
    \subfigure[Samples with lowest $p(y|\bs{x})$]{
        \includegraphics[width=0.48\columnwidth]{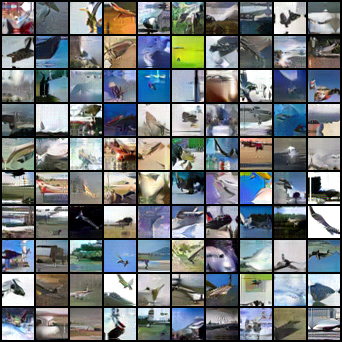}
    }
    \caption{JEM++ generated class-conditional samples of \textbf{Plane}}
    \label{figure:class_0}
\end{figure*}

\begin{figure*}[ht!]
    \centering
    \subfigure[Samples with highest $p(\bs{x}$)]{
        \includegraphics[width=0.48\columnwidth]{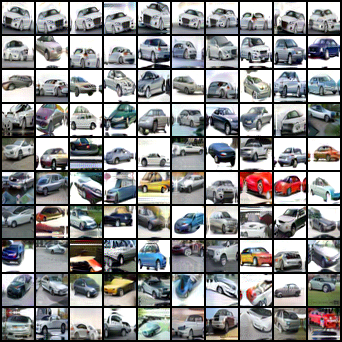}
    }
    \subfigure[Samples with lowest $p(\bs{x})$]{
        \includegraphics[width=0.48\columnwidth]{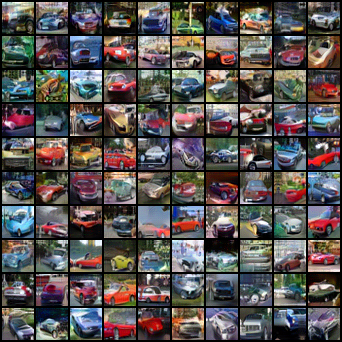}
    }
    \subfigure[Samples with highest $p(y|\bs{x}$)]{
        \includegraphics[width=0.48\columnwidth]{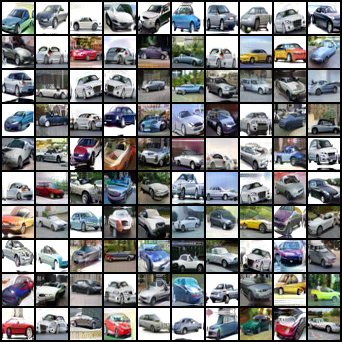}
    }
    \subfigure[Samples with lowest $p(y|\bs{x})$]{
        \includegraphics[width=0.48\columnwidth]{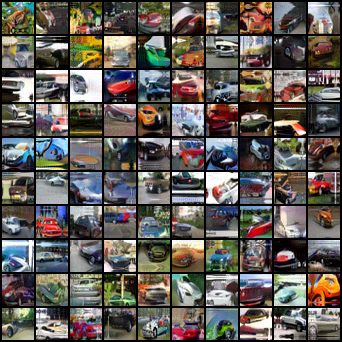}
    }
    \caption{JEM++ generated class-conditional samples of \textbf{Car}}
    \label{figure:class_1}
\end{figure*}

\begin{figure*}[ht!]
    \centering
    \subfigure[Samples with highest $p(\bs{x}$)]{
        \includegraphics[width=0.48\columnwidth]{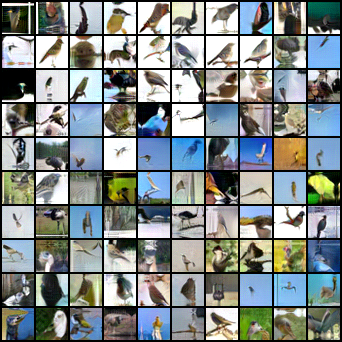}
    }
    \subfigure[Samples with lowest $p(\bs{x})$]{
        \includegraphics[width=0.48\columnwidth]{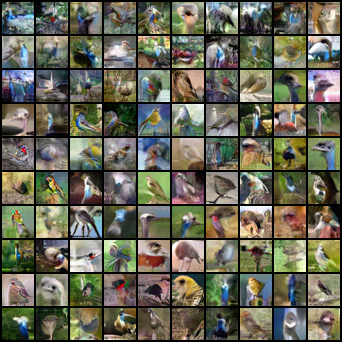}
    }
    \subfigure[Samples with highest $p(y|\bs{x}$)]{
        \includegraphics[width=0.48\columnwidth]{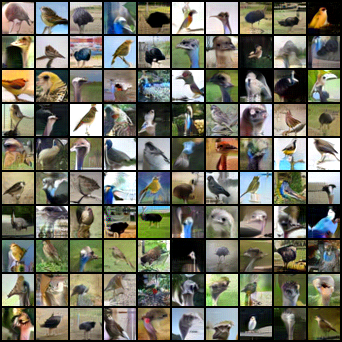}
    }
    \subfigure[Samples with lowest $p(y|\bs{x})$]{
        \includegraphics[width=0.48\columnwidth]{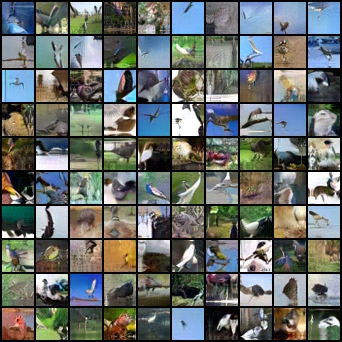}
    }
    \caption{JEM++ generated class-conditional samples of \textbf{Bird}}
    \label{figure:class_2}
\end{figure*}

\begin{figure*}[ht!]
    \centering
    \subfigure[Samples with highest $p(\bs{x}$)]{
        \includegraphics[width=0.48\columnwidth]{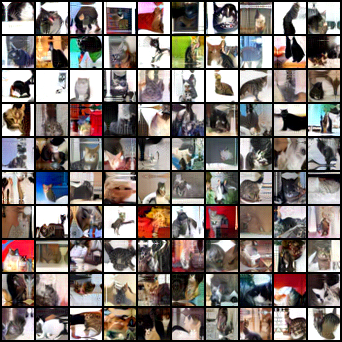}
    }
    \subfigure[Samples with lowest $p(\bs{x})$]{
        \includegraphics[width=0.48\columnwidth]{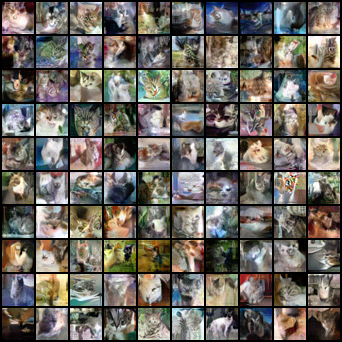}
    }
    \subfigure[Samples with highest $p(y|\bs{x}$)]{
        \includegraphics[width=0.48\columnwidth]{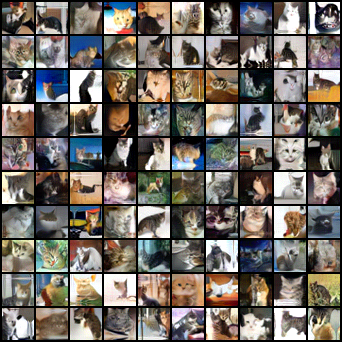}
    }
    \subfigure[Samples with lowest $p(y|\bs{x})$]{
        \includegraphics[width=0.48\columnwidth]{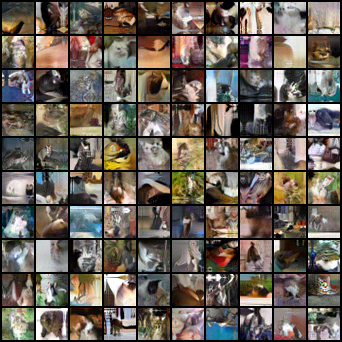}
    }
    \caption{JEM++ generated class-conditional samples of \textbf{Cat}}
    \label{figure:class_3}
\end{figure*}

\begin{figure*}[ht!]
    \centering
    \subfigure[Samples with highest $p(\bs{x}$)]{
        \includegraphics[width=0.48\columnwidth]{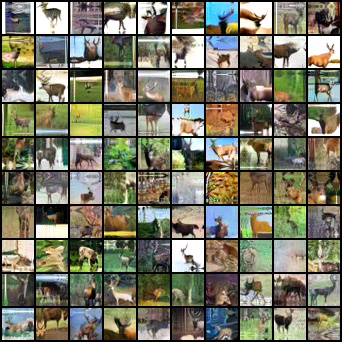}
    }
    \subfigure[Samples with lowest $p(\bs{x})$]{
        \includegraphics[width=0.48\columnwidth]{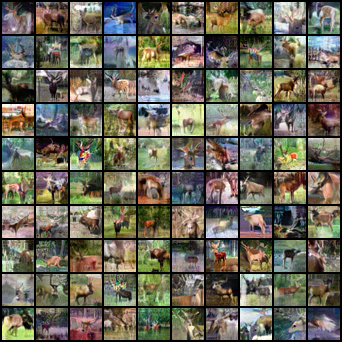}
    }
    \subfigure[Samples with highest $p(y|\bs{x}$)]{
        \includegraphics[width=0.48\columnwidth]{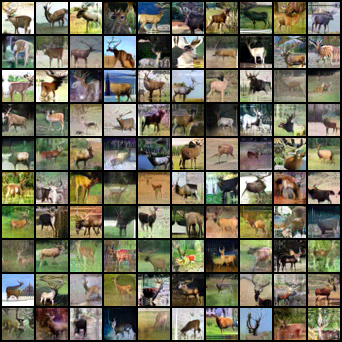}
    }
    \subfigure[Samples with lowest $p(y|\bs{x})$]{
        \includegraphics[width=0.48\columnwidth]{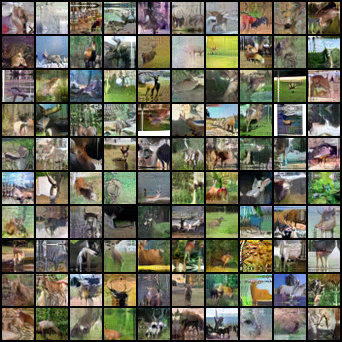}
    }
    \caption{JEM++ generated class-conditional samples of \textbf{Deer}}
    \label{figure:class_4}
\end{figure*}

\begin{figure*}[ht!]
    \centering
    \subfigure[Samples with highest $p(\bs{x}$)]{
        \includegraphics[width=0.48\columnwidth]{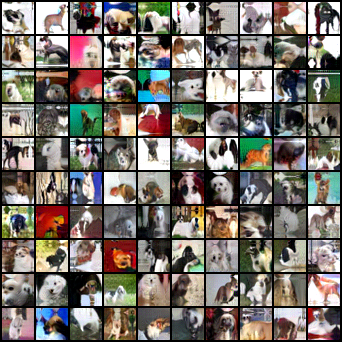}
    }
    \subfigure[Samples with lowest $p(\bs{x})$]{
        \includegraphics[width=0.48\columnwidth]{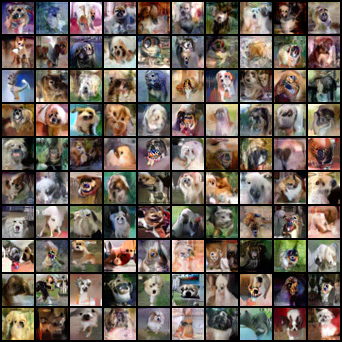}
    }
    \subfigure[Samples with highest $p(y|\bs{x}$)]{
        \includegraphics[width=0.48\columnwidth]{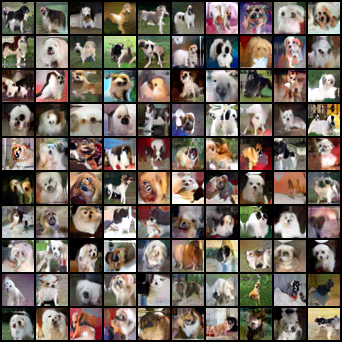}
    }
    \subfigure[Samples with lowest $p(y|\bs{x})$]{
        \includegraphics[width=0.48\columnwidth]{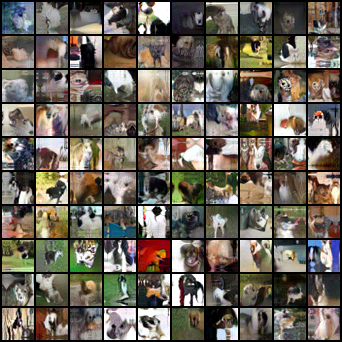}
    }
    \caption{JEM++ generated class-conditional samples of \textbf{Dog}}
    \label{figure:class_5}
\end{figure*}

\begin{figure*}[ht!]
    \centering
    \subfigure[Samples with highest $p(\bs{x}$)]{
        \includegraphics[width=0.48\columnwidth]{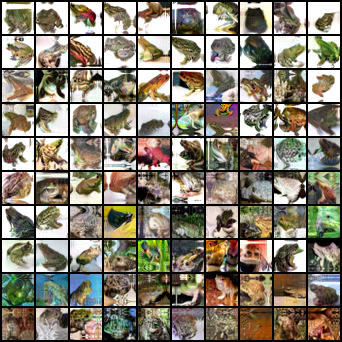}
    }
    \subfigure[Samples with lowest $p(\bs{x})$]{
        \includegraphics[width=0.48\columnwidth]{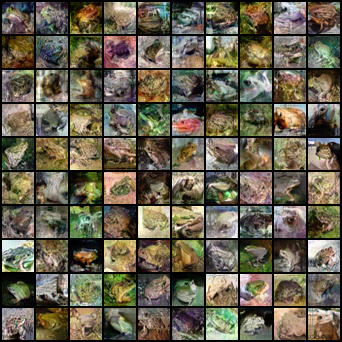}
    }
    \subfigure[Samples with highest $p(y|\bs{x}$)]{
        \includegraphics[width=0.48\columnwidth]{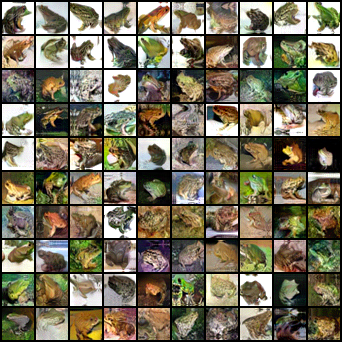}
    }
    \subfigure[Samples with lowest $p(y|\bs{x})$]{
        \includegraphics[width=0.48\columnwidth]{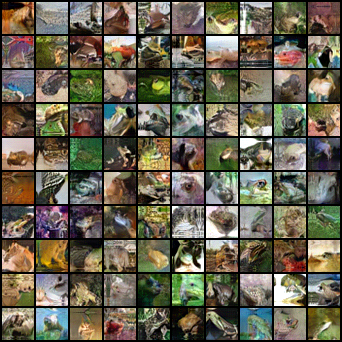}
    }
    \caption{JEM++ generated class-conditional samples of \textbf{Frog}}
    \label{figure:class_6}
\end{figure*}

\begin{figure*}[ht!]
    \centering
    \subfigure[Samples with highest $p(\bs{x}$)]{
        \includegraphics[width=0.48\columnwidth]{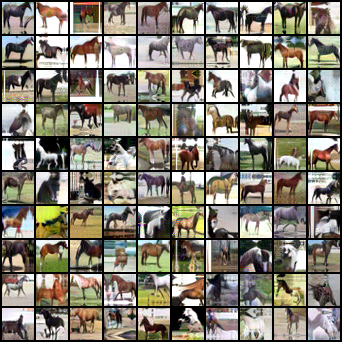}
    }
    \subfigure[Samples with lowest $p(\bs{x})$]{
        \includegraphics[width=0.48\columnwidth]{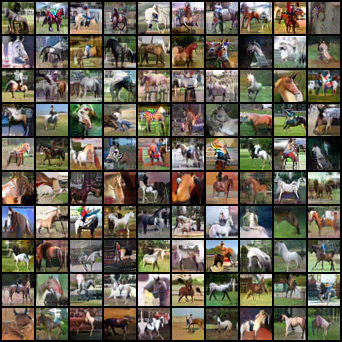}
    }
    \subfigure[Samples with highest $p(y|\bs{x}$)]{
        \includegraphics[width=0.48\columnwidth]{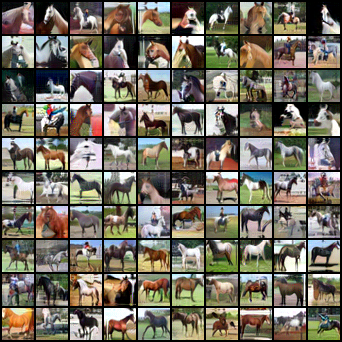}
    }
    \subfigure[Samples with lowest $p(y|\bs{x})$]{
        \includegraphics[width=0.48\columnwidth]{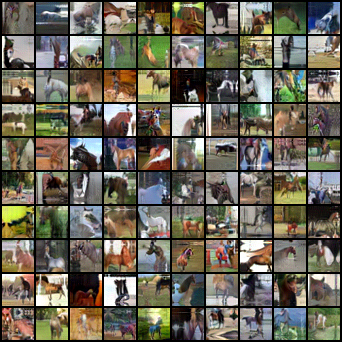}
    }
    \caption{JEM++ generated class-conditional samples of \textbf{Horse}}
    \label{figure:class_7}
\end{figure*}

\begin{figure*}[ht!]
    \centering
    \subfigure[Samples with highest $p(\bs{x}$)]{
        \includegraphics[width=0.48\columnwidth]{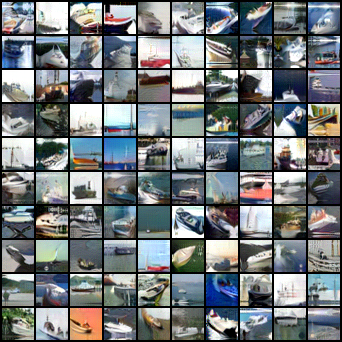}
    }
    \subfigure[Samples with lowest $p(\bs{x})$]{
        \includegraphics[width=0.48\columnwidth]{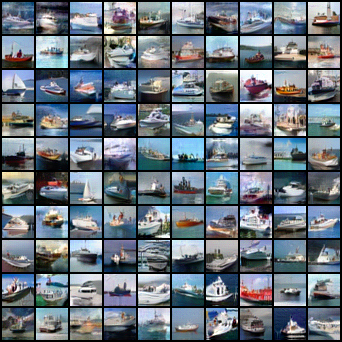}
    }
    \subfigure[Samples with highest $p(y|\bs{x}$)]{
        \includegraphics[width=0.48\columnwidth]{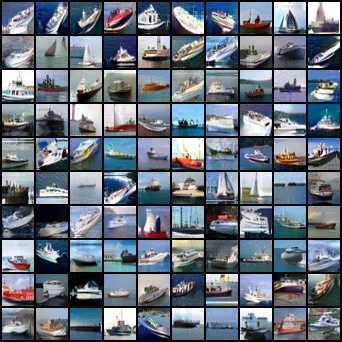}
    }
    \subfigure[Samples with lowest $p(y|\bs{x})$]{
        \includegraphics[width=0.48\columnwidth]{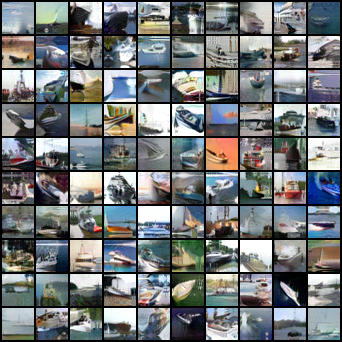}
    }
    \caption{JEM++ generated class-conditional samples of \textbf{Ship}}
    \label{figure:class_8}
\end{figure*}

\begin{figure*}[ht!]
    \centering
    \subfigure[Samples with highest $p(\bs{x}$)]{
        \includegraphics[width=0.48\columnwidth]{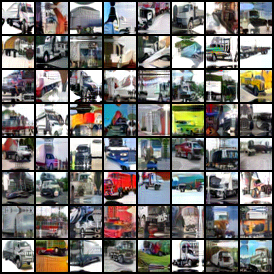}
    }
    \subfigure[Samples with lowest $p(\bs{x})$]{
        \includegraphics[width=0.48\columnwidth]{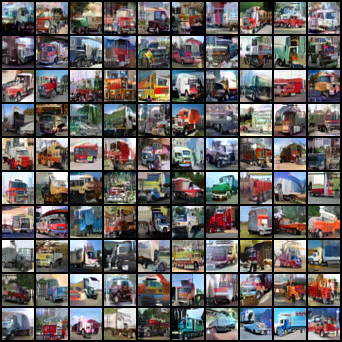}
    }
    \subfigure[Samples with highest $p(y|\bs{x}$)]{
        \includegraphics[width=0.48\columnwidth]{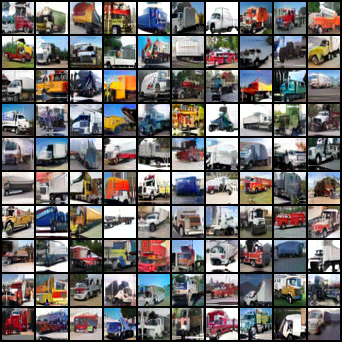}
    }
    \subfigure[Samples with lowest $p(y|\bs{x})$]{
        \includegraphics[width=0.48\columnwidth]{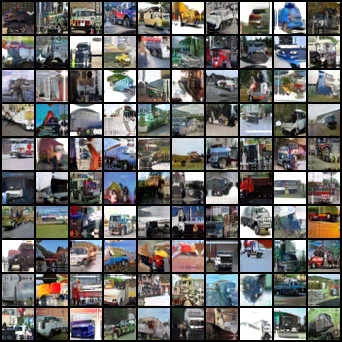}
    }
    \caption{JEM++ generated class-conditional samples of \textbf{Truck}}
    \label{figure:class_9}
\end{figure*}

Additional JEM++ generated samples of SVHN and CIFAR100 are provided in Figure~\ref{figure:svhn_CIFAR100_cond}. Additional JEM++ generated class-conditional (best and worst) samples of CIFAR10 are provided in Figures~\ref{figure:class_0}-\ref{figure:class_9}.  It is worth noting that the worst images (the lowest $p(\bs{x})$ or $p(y|\bs{x})$) generated by JEM++ are more visually appealing than JEM generated (see examples in the Appendix of JEM~\cite{jem}).

\end{document}